%% file: main.tex
\documentclass[10pt,twocolumn,letterpaper]{article}

\usepackage{cvpr}              

\usepackage{url}            
\usepackage{booktabs}       
\usepackage{amsfonts}       
\usepackage{nicefrac}       
\usepackage{microtype}      
\usepackage{algorithm}
\usepackage{algorithmic}
\usepackage{graphicx}
\usepackage{makecell}
\usepackage{multirow}
\usepackage{multicol}
\usepackage{bm}
\usepackage{booktabs} 
\usepackage{overpic}
\usepackage{pict2e}
\usepackage{amsmath}
\usepackage{xcolor}

\usepackage{listings}
\usepackage{anyfontsize}
\usepackage{diagbox}
\usepackage{textcomp}
\usepackage{rotating}
\usepackage{amssymb}
\usepackage{pifont}

\usepackage{colortbl}
\newcommand\algcomment[1]{\def\@algcomment{\footnotesize#1}}

\usepackage[pagebackref,breaklinks,colorlinks]{hyperref}

\usepackage[capitalize]{cleveref}
\crefname{section}{Sec.}{Secs.}
\Crefname{section}{Section}{Sections}
\Crefname{table}{Table}{Tables}
\crefname{table}{Tab.}{Tabs.}

\definecolor{myblue}{RGB}{169,196,235}
\definecolor{mygreen}{RGB}{213,232,212}
\definecolor{mygray}{RGB}{191,191,191}

\graphicspath{{figures/}{figures/pipeline/}}
\newcommand{\myPara}[1]{\vspace{-.08in}\paragraph{#1}}

\ifdefined \GramaCheck
\newcommand{\CheckRmv}[1]{}
\newcommand{\figref}[1]{Figure 1}%
\newcommand{\tabref}[1]{Table 1}%
\newcommand{\secref}[1]{Section 1}
\renewcommand{\eqref}[1]{Equation 1}
\else
\newcommand{\CheckRmv}[1]{#1}
\newcommand{\figref}[1]{Fig.~\ref{#1}}%
\newcommand{\tabref}[1]{Tab.~\ref{#1}}%
\newcommand{\secref}[1]{Sec.~\ref{#1}}
\newcommand{\algref}[1]{Alg.~\ref{#1}}
\renewcommand{\eqref}[1]{Eqn.~(\ref{#1})}
\fi
\def\vs{\emph{vs.}}
\def\ie{\emph{i.e.,~}}
\def\eg{\emph{e.g.,~}}
\def\etc{\emph{etc}}

\def\cvprPaperID{1352} 
\def\confName{CVPR}
\def\confYear{2022}

\begin{document}

\title{Representation Compensation Networks for Continual Semantic Segmentation}

\author{
Chang-Bin Zhang $^{1*}$ \qquad
Jia-Wen Xiao $^1$\thanks{The first two authors contribute equally.} \qquad
Xialei Liu $^1$\thanks{Corresponding author (xialei@nankai.edu.cn)} \qquad
Ying-Cong Chen $^2$ \qquad
Ming-Ming Cheng $^1$ \\
$^1$ TKLNDST, CS, Nankai University \qquad
$^2$ The Hongkong University of Science and Technology\\
}

\maketitle

\begin{abstract}
In this work, we study the continual semantic segmentation problem, where the deep neural networks are required to incorporate new classes continually without catastrophic forgetting. 
We propose to use a structural re-parameterization mechanism, named representation compensation (RC) module,
to decouple the representation learning of both old and new knowledge. The RC module consists of two dynamically evolved branches with one frozen and one trainable.
Besides, we design a pooled cube knowledge distillation strategy 
on both spatial and channel dimensions 
to further enhance the plasticity and stability of the model.
We conduct experiments on two challenging continual semantic segmentation scenarios,
continual class segmentation and continual domain segmentation.
Without any extra computational overhead and parameters during inference, 
our method outperforms state-of-the-art performance.
The code is available at \url{https://github.com/zhangchbin/RCIL}.
\end{abstract}

\section{Introduction}
Data-driven deep neural networks
\cite{seyedhosseini2016semantic,zand2016ontology,nirkin2021hyperseg,zhu2021learning}
have made many milestones in semantic segmentation.
However, these fully-supervised models \cite{ding2020semantic,yang2021is,chen2021spatial}
can only handle a fixed number of classes.
In real-world applications,
it is preferable that a model can be dynamically extended to identify new classes.
A straightforward solution is to rebuild the training set and retrain the model with all data available, known as \emph{Joint Training}.
However, considering the cost of retraining models, 
sustainable development of algorithms and privacy issues, 
it is particularly crucial to update the model with only current data 
to achieve the goal of recognizing both new and old classes.
Nevertheless,
naively fine-tuning a trained model with new data can result in catastrophic forgetting~\cite{kirkpatrick2017overcoming}.
Therefore,
in this paper,
we seek continual learning,
which can potentially enable a model to recognize new categories without catastrophic forgetting.

\begin{figure}[!tp] 
  \centering
  \small
  \begin{overpic}[width=.95\linewidth]{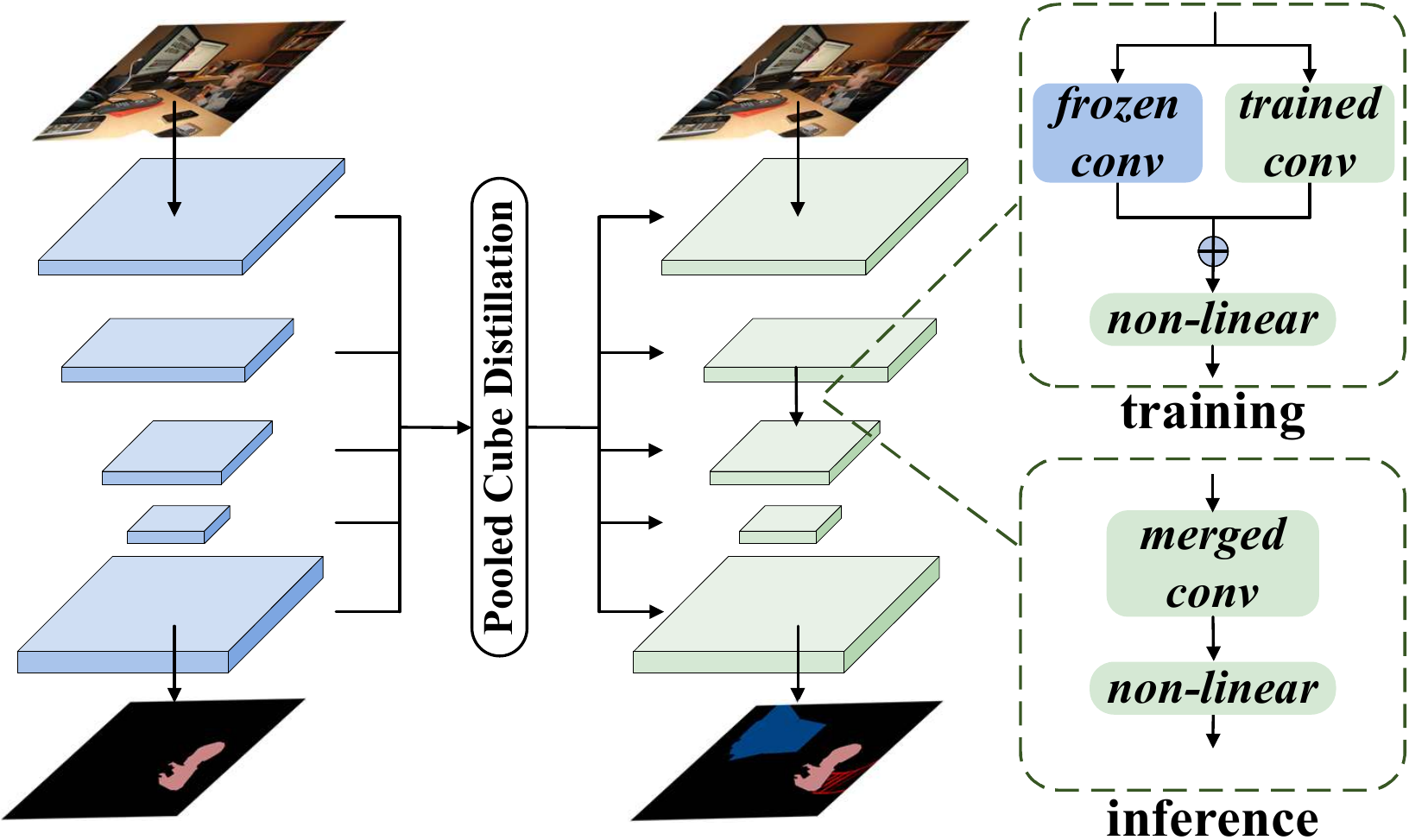}
            \put(77.2, -4.5) {\small{ RC Module}}
		    \put(7, -4) {\small{$f_{t-1}$}}
		    \put(53.5, -4) {\small{$f_{t}$}}
  \end{overpic}
  \vskip .1in
  \caption{Illustration of our proposed training framework for 
    continual semantic segmentation to avoid catastrophic forgetting.
    We design two mechanisms in our method, 
    representation compensation (RC) module 
    and pooled cube distillation (PCD).
  }\label{fig:framework}
\end{figure}

In the scenario of continual semantic segmentation~\cite{ilt,mib,plop,sdr},
given the previously trained model and the training data of the new classes,
the model is supposed to distinguish all seen classes,
including previous classes (old classes) and new classes.
However, to save the labeling cost,
the new training data often only has labels for the new classes,
treating old classes as background.
Learning with the new data directly without any additional designs is very challenging, which can easily lead to
catastrophic forgetting~\cite{kirkpatrick2017overcoming}.

As indicated in~\cite{li2017learning,kirkpatrick2017overcoming,douillard2020podnet},
fine-tuning the model on new data may lead to catastrophic forgetting, \ie{the model quickly fits the data distribution of the new classes, while losing the discrimination for the old classes}.
Some methods~\cite{kirkpatrick2017overcoming,pan2020continual,iscen2020memory,liu2020more,tao2020topology,yu2020semantic,park2019continual} 
play regularization on model parameters to improve its stability.
However, all parameters are updated on the training data of
the new classes.
This is however challenging,
as new and old knowledge are entangled together in model parameters,
making it extremely difficult to keep the fragile balance of learning new knowledge and keeping old ones.
Some other methods \cite{Liu2020AANets,verma2021efficient,singh2021rectification,yan2021dynamically,singh2020calibrating,kanakis2020reparameterizing}
increase the capacity of the model to have a better trade-off of stability 
and plasticity, but with the cost of growing memory of the network.

In this work, we propose an easy-to-use representation compensation module,
aiming at remembering the old knowledge
while allowing extra capacity for new knowledge.
Inspired by structural re-parameterization~\cite{ding2019acnet, ding2021repvgg},
we replace the convolution layers in the network with two parallel branches during training,
which is named as representation compensation module.
As shown in~\figref{fig:framework}, during training,
the output of two parallel convolutions is fused before the non-linear activation layer.
At the beginning of  each continual learning step,
we equivalently merge the parameters
of the two parallel convolutions into one convolution,
which will be frozen to retain the old knowledge.
Another branch is trainable and it inherits the parameters from
the corresponding branch in the previous step.
The representation compensation strategy is supposed to 
remember the old knowledge using the frozen branch
while allowing extra capacity for new knowledge using the trainable branch.
Importantly, this module brings no extra parameters and computation cost during inference.

To further alleviate catastrophic forgetting~\cite{kirkpatrick2017overcoming},
we introduce a knowledge distillation mechanism~\cite{romero2014fitnets} between intermediate layers (shown in~\figref{fig:framework}),
named Pooled Cube Distillation. It can suppress the negative impact of errors and noises in local feature maps.
The main contributions of this paper are:

\begin{itemize}
\item We propose a representation compensation module with 
two branches during training, 
one for retaining the old knowledge and 
one for adapting to new data. 
It always keeps the same computation and memory cost during inference 
as the number of tasks grows.

\item We conduct experiments on \emph{continual class segmentation} and
\emph{continual domain segmentation}, respectively.
Experimental results demonstrate that our method outperforms the state-of-the-art performance on three different datasets.

\end{itemize}

\section{Related Work}

\myPara{Semantic Segmentation.}
Early methods focused on modeling contextual relationships
\cite{koltun2011efficient,zheng2015conditional,arnab2016higher}.
Currently methods pay more attention to multi-scale feature aggregation
\cite{long2015fully,hariharan2015hypercolumns,noh2015learning,lin2018multi,lin2017refinenet,badrinarayanan2017segnet,peng2017large,tian2019decoders}.
Some methods~\cite{liu2017learning,li2019expectation,ding2018context,chen2016attention,hong2016learning,fu2019dual,hou2020strip}
is inspired by Non-local~\cite{wang2018non}, 
utilizing attention mechanisms to establish connections between image contexts.
Another line of research~\cite{yang2018denseaspp,chen2018encoder,
mehta2018espnet} aimed at fusing features from different receptive fields.
Recently, transformer architectures
\cite{carion2020end,zeng2020learning,dosovitskiy2020image,zhu2020deformable,wang2020end,zheng2020rethinking} 
shine in semantic segmentation,
focusing on multi-scale feature fusion~\cite{xie2021segformer, chen2021crossvit, wang2021pyramid, zhang2020feature}
and contextual feature aggregation~\cite{strudel2021segmenter, liu2021swin}.

\begin{figure*}[!t] 
  \begin{overpic}[width=1.0\textwidth,tics=2]{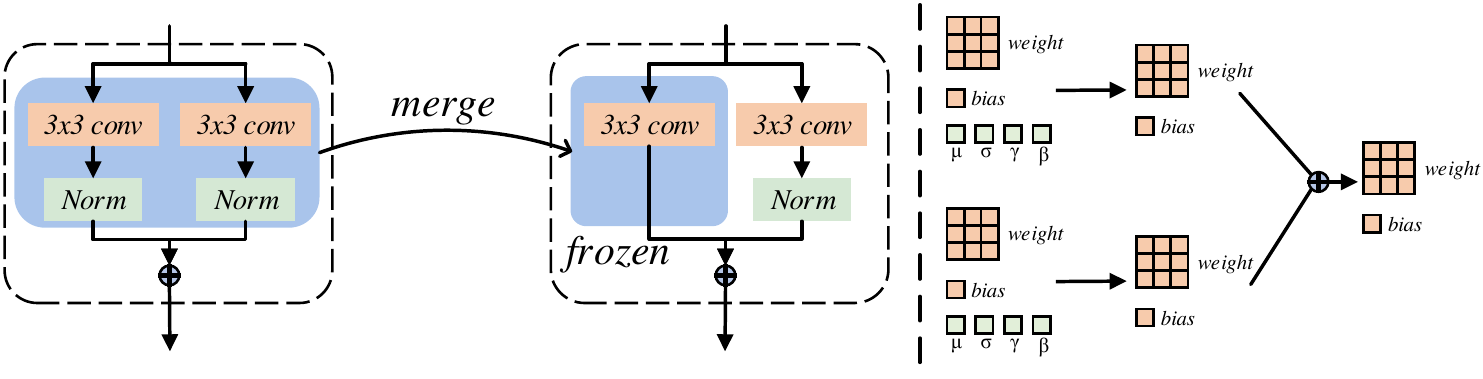}
    \put(8, 0){step $t-1$}
    \put(47, 0){step $t$}
  \end{overpic}
  \vskip -0.1in
  \caption{Illustration of our representation compensation mechanism.
	We modify the $3\times3$ convolution as two parallel convolutions.
	The features from the two branches are aggregated before the activation layer.
	At the beginning of step $t$, thus, the two parallel branches trained at step $t-1$ can be merged into an equivalent convolution layer, which will be frozen and is regarded as one branch of $step~t$.
	Another branch in $step~t$ is initialized from the corresponding branch from $step~t-1$.
	We demonstrate the merge operation in the right part of the figure.
  }\label{fig:method}
  \vskip -0.2in
\end{figure*}

\myPara{Continual Learning.}
Continual learning focuses on alleviating catastrophic forgetting
while being discriminative for newly learned classes.
To solve this problem,
many work~\cite{kim2021continual,bang2021rainbow,chaudhry2020using,smith2021always,belouadah2019il2m}
propose to review knowledge by rehearsal-based mechanism.
The knowledge can be stored by multiple types,
like examples~\cite{verwimp2021rehearsal,cha2021co2l,bang2021rainbow,shim2020online,chaudhry2020using,buzzega2020dark},
prototypes~\cite{zhu2021prototype,zhu2021self,hayes2020remind},
generative networks~\cite{maracani2021recall}, \etc.
Although these rehearsal-based methods usually achieve high performance,
they need storage and authority for storing.
In the more challenging scenario without any replay,
many methods explore regularization to maintain old knowledge,
including knowledge distillation~\cite{chaudhry2018riemannian,li2017learning,rebuffi2017icarl,simon2021learning,cheraghian2021semantic,douillard2020podnet,dhar2019learning},
adversarial training~\cite{ebrahimi2020adversarial,xiang2019incremental},
vanilla regularization~\cite{zenke2017continual,kirkpatrick2017overcoming,pan2020continual,iscen2020memory,liu2020more,tao2020topology,yu2020semantic,park2019continual} and so on.
Others focus on the capacity of the neural network.
One of the research line~\cite{Liu2020AANets,verma2021efficient,singh2021rectification,yan2021dynamically,singh2020calibrating,kanakis2020reparameterizing} is to expand the network architecture while learning new knowledge.
Another research line~\cite{jung2020continual,abati2020conditional} explores the sparsity regularization for network parameters,
which aims at activating as few neurons as possible for each task.
This sparsity regularization reduces the redundancy in the network,
while limiting the learning capacity for each task.
Some work propose to learn better representations by combining self-supervised learning for feature extractor~\cite{wu2021striking,cha2021co2l}
and solving class imbalance~\cite{zhang2021few,liu2020incremental,kim2020imbalanced,zhao2020maintaining,hou2019learning}.

\myPara{Continual Semantic Segmentation.}
Continual semantic segmentation is still an urgent problem to solve, mainly focusing on
catastrophic forgetting\cite{kirkpatrick2017overcoming} in semantic segmentation.
In this field, continual class segmentation is a classic setting, with great progress made by several previous work: \cite{yan2021em,huang2021half} explore rehearsal-based methods to review old knowledge;
MiB~\cite{mib} models the potential classes to solve the ambiguous of \emph{background} class;
PLOP~\cite{plop} applies knowledge distillation strategy to intermediate layers;
SDR~\cite{sdr} takes advantage of prototype matching to perform consistency constraints in the latent space representation. While others ~\cite{stan2020unsupervised, frey2021continual, yu2020semantic} utilize high-dimensional information, self-training and model adaptation to overcome this problem.
Moreover, continual domain segmentation is a novel setting proposed by PLOP~\cite{plop}, aiming at integrating new domain rather than new classes.
Different from previous methods,
we focus on expanding the network dynamically,
decoupling the representation learning of old classes and new classes.

\section{Method}
\subsection{Preliminaries} \label{sec:prelimi}
Let $\mathcal{D}=\{x_i, y_i\}$ denotes the training set,
where $x_i$ denotes the input image and $y_i$ is the corresponding segmentation ground-truth.
In the challenging continual learning scenario,
we call each training on the newly added dataset $\mathcal{D}_t$ as a \emph{step}.
At step $t$,
given a model $f_{t-1}$ with parameter $\theta_{t-1}$ trained on $\{\mathcal{D}_0,\mathcal{D}_1...\mathcal{D}_{t-1} \}$ with $\{C_0, C_1...C_{t-1}\}$ classes continually, the model is supposed to learn the discrimination for $\sum_{n=0}^t C_n$ classes when it encounters a newly added dataset $\mathcal{D}_t$ with extra $C_t$ new classes.
When training on $\mathcal{D}_t$, the training data of old classes are not accessible. Besides, to save the training cost, the ground-truth in $\mathcal{D}_t$ only contains the $C_t$ new classes, while the old classes are labeled as \emph{background}.
Thus,
there is an urgent problem,
catastrophic forgetting.
To verify the effectiveness of different methods,
it is often necessary to perform the continual learning multiple times
\eg{$N$ steps}.

\subsection{Representation Compensation Networks}\label{sec:rcnet}

To decouple the retaining of old knowledge and learning of new knowledge, as shown in~\figref{fig:method},
we introduce our representation compensation mechanism.
In most of the deep neural networks,
a $3\times 3$ convolution followed by normalization and non-linear activation layer is a common component.
We modify this architecture by adding a parallel $3\times 3$ convolution followed by a normalization layer for each component.
The output of two parallel convolution-normalization
layers is fused,
then is rectified by a non-linear activation layer. 
Formally, this architecture contains two parallel convolution layers
with weight $\{W^0, W^1\}$ and bias $\{b^0, b^1\}$, followed by two independent normalization layers, respectively.
Let $Norm^0 = \{\mu^0, \sigma^0, \gamma^0, \beta^0 \}$
and $Norm^1 = \{\mu^1, \sigma^1, \gamma^1, \beta^1 \}$
denote the mean, variance, weight and bias of two normalization layers $Norm^0$ and $Norm^1$.
Thus,
the calculation of input $x$ before non-linear activation function can be denoted as  
\begin{equation}
\begin{aligned}
\hat{x} &=\sum_{i=0}^1 Norm_i(W_i x + b_i)\\
&=\sum_{i=0}^1 (\gamma_i \frac{W_i x + b_i - \mu_i}{\sigma_i} + \beta_i) \\
&= (\sum_{i=0}^1\frac{\gamma_i W_i}{\sigma_i} )x + \sum_{i=0}^1(\frac{\gamma_i b_i - \gamma_i \mu_i}{\sigma_i}
 + \beta_i)  \\
&= \hat{W} x + \hat{b}.
\end{aligned}
\end{equation}
This equation demonstrates that two parallel branches can be equivalently represented as one with weight $\hat{W}$ and bias $\hat{b}$.
We also display the transformation in the right part of~\figref{fig:method}.
Therefore,
for this modified architecture,
we can equivalently merge the parameters of two branches into
one convolution.

More precisely, in step $0$,
all parameters are trainable to train a model that can discriminate $\mathcal{C}_0$ classes.
For the subsequent learning steps,
the model is supposed to segment newly added classes.
In these continual learning steps,
the network will be initialized with the parameters trained
in the previous step,
which is beneficial to transfer knowledge~\cite{mib}.
At the beginning of step $t$,
since the model is supposed to avoid forgetting old knowledge,
we merge the parallel branches trained in step $t-1$ to one convolution layer.
The parameters in this merged branch are frozen to memorize
the old knowledge,
as shown in~\figref{fig:method}.
Another branch is trainable to learn new knowledge,
which is initialized with the corresponding
branch in the previous step.
Besides,
we design a drop-path strategy,
which is applied on aggregating the output, $x_1$  and $x_2$ from two branches.
During training,
the output before the non-linear activation is denoted as 
\begin{equation}
    \hat{x} = \eta \cdot x_1 + (1 - \eta) \cdot x_2,
\end{equation}
where $\eta$ is the random channel-wise weighted vector and sampling from the set $\{0, 0.5, 1\}$ uniformly.
During inference,
the element of vector $\eta$ is set as $0.5$.
Experimental results demonstrate that this strategy brings slight improvement.

\begin{figure}[t]
    \centering
	\begin{small}
		\begin{overpic}[width=\linewidth,tics=8]{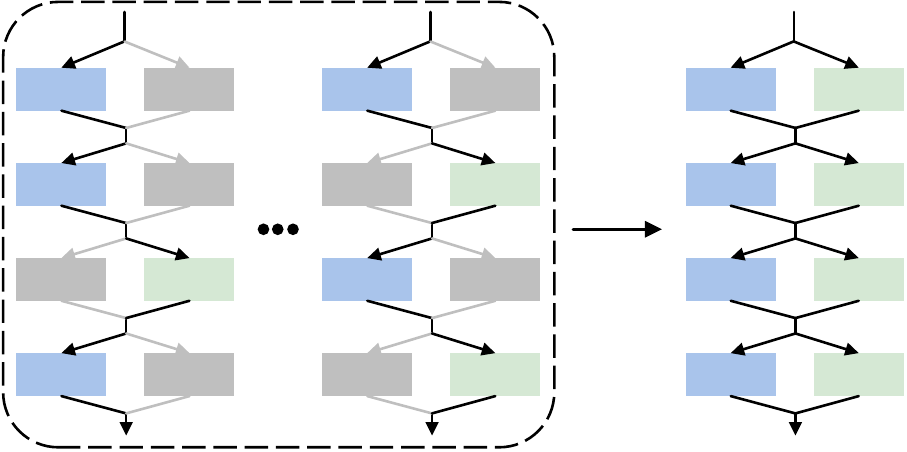}
		    \put(62,27) {\footnotesize{Ensemble}}
		    \put(12, -3.5) {(a)}
		    \put(46, -3.5) {(b)}
		    \put(86, -3.5) {(c)}
		\end{overpic}
	\end{small}
	\vskip -0.1in
	\caption{Illustration of our proposed Representation Compensation Network. Our architecture (c) can be regarded as an implicit ensemble of numerous sub-networks (a), (b), \etc. The \colorbox{myblue}{blue} denotes the frozen layers inherited from the merged teacher model. The \colorbox{mygreen}{green} denotes the trainable layers. The \colorbox{mygray}{gray} denotes the layers that are ignored in the sub-network.
	}\label{fig:subnetwork}
\end{figure}

\textbf{Analysis on RC-Module's Effectiveness.}
As shown in~\figref{fig:subnetwork},
the parallel convolution structure can be regarded as
an implicit ensemble~\cite{huang2016deep,he2016deep} of numerous sub-networks.
The parameters of some layers in these sub-networks are inherited
from the the merged teacher model (trained at previous step) and are frozen.
During training,
similar to~\cite{xu2020bert,fu2020interactive},
these frozen teacher layers will impose regularization to trainable parameters,
encouraging trainable layers to behave like the teacher model.
In a special case where only one layer in the sub-network is trainable, as shown in~\figref{fig:subnetwork}(a),
during training,
this layer will take into account both adapting for the representation of frozen layers and learning for new knowledge.
Therefore,
this mechanism will alleviate catastrophic forgetting of the trainable layer.
We further promote this effect to general sub-networks like~\figref{fig:subnetwork}(b),
which will also encourage the trainable layers to adapt to the representation of the frozen layers.
Furthermore,
all sub-networks are ensembled,
integrating knowledge from different sub-networks to one network, like~\figref{fig:subnetwork}(c).

\subsection{Pooled Cube Knowledge Distillation}

\begin{figure}[t]
    \centering
	\begin{small}
		\begin{overpic}[width=0.45\textwidth,tics=8]{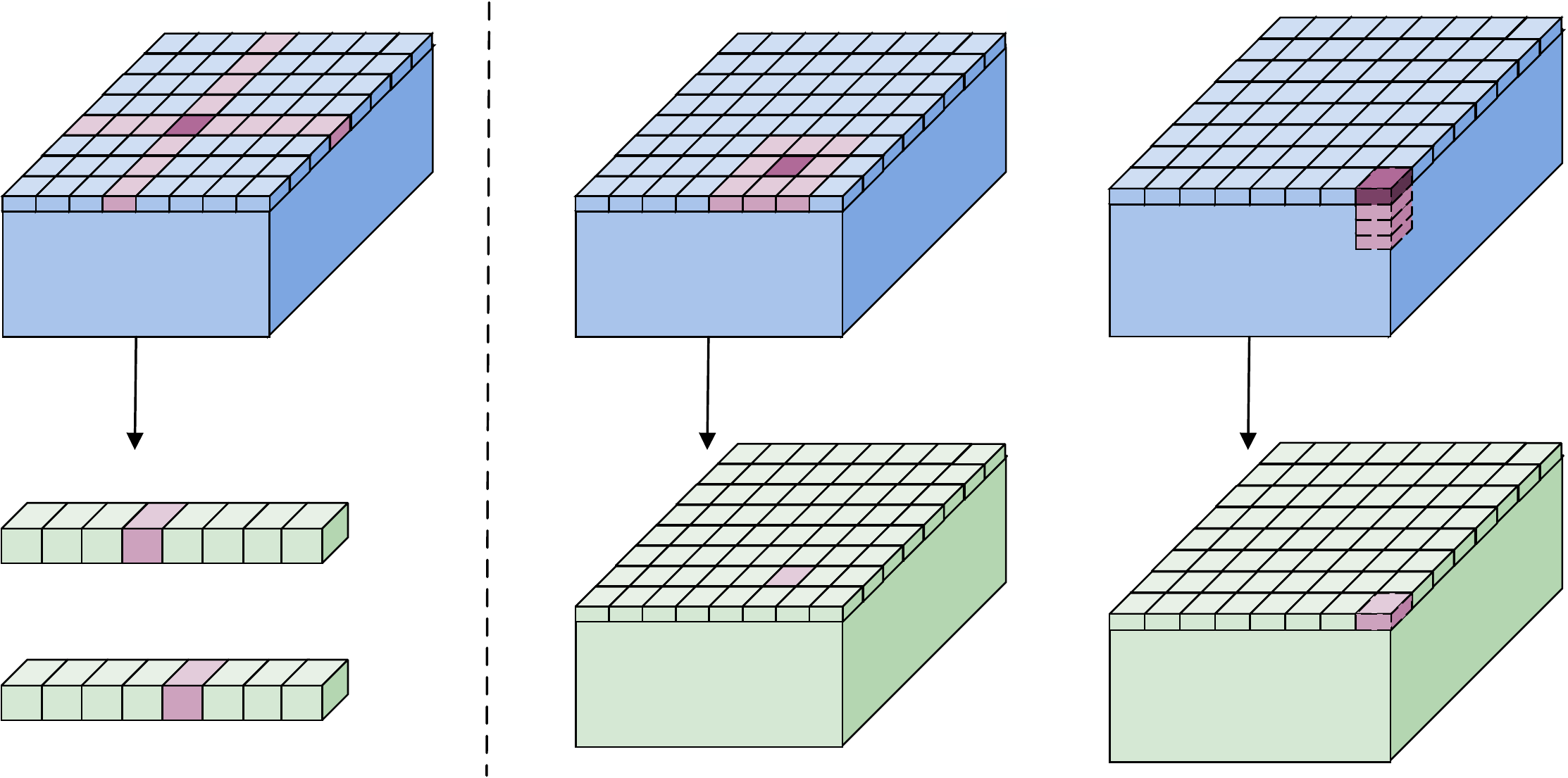}
		    \put(9, 24){strip pooling}
		    \put(46,24){avg. pooling}
		    \put(80.2,24){avg. pooling}
		    \put(8,10.2){row}
		    \put(5,0){column}
		    \put(42,-3) {spatial KD}
		    \put(72,-3){Channel-wise KD}
		    \put(3,-7){(a) PLOP~\cite{plop}}
		    \put(60,-7){(b) Ours}
		\end{overpic}
	\end{small}
	\vskip 0.15in
	\caption{Comparison between PLOP~\cite{plop} and our proposed Pooled Cube Knowledge Distillation mechanism.
	}\label{fig:kd}
\end{figure}

In order to further alleviate the forgetting of old knowledge,
following PLOP~\cite{plop}, we also explore  distilling knowledge between intermediate layers.
As shown in~\figref{fig:kd}(a),
PLOP~\cite{plop} introduces strip pooling~\cite{hou2020strip} to integrate features from the teacher model and current model, respectively.
The pooling operation plays a key role in keeping discrimination for old classes and allowing learning new classes.
In our method,
we design the average pooling-based knowledge distillation along the spatial dimension.
Additionally, we use the average pooling in the channel dimension at each position as well to maintain their individual activation intensity.
Overall, as shown in~\figref{fig:kd}(b),
we use the average pooling on both spatial and channel dimensions.

Formally,
we select feature maps $\{X^1, X^2, ..., X^L\}$ before the last non-linear activation layer for all $L$ stages,
including decoder and all stages in the backbone.
For the features from the teacher model and the student model,
we firstly calculate the square of value at each pixel to retain the negative information.
Then,
we perform multi-scale average pooling on spatial
and channel dimensions, respectively.
The features $\hat{X_T^l}, \hat{X_S^l}$ of the teacher model and the student model can be calculated by the average pooling operation $\odot$:
\begin{equation}
\begin{aligned}
\hat{X}_T^{l, m} &= M \odot [(X_{T,ij}^l)^2]\\
\hat{X}_S^{l, m} &= M \odot [(X_{S,ij}^l)^2],
\end{aligned}
\end{equation}
where $M$ denotes the $m_{th}$ average pooling kernel,
and $l$ denotes the $l_{th}$ stage.
For the average pooling on the spatial dimension,
we use the multi-scale windows to model the relationships between pixels in the local region.
The size of kernel $M$ belongs to $\mathcal{M}=\{4,8,12,16,20,24\}$ and the step size is set to 1.
And we simply set the window size as $3$ for the average pooling on channel dimension.
Then,
the spatial knowledge distillation loss function $L_{skd}$ for the intermediate layers can be denoted as
\begin{equation}
\begin{aligned}
L_{skd} = \frac{1}{L}\frac{1}{|\mathcal{M}|}\sum_{l=1}^{L}\sum_{m=1}^{|\mathcal{M}|} \sqrt{\sum_{i=1}^{H} \sum_{j=1}^{W} \sum_{d=1}^{D} [(\hat{X}_{T, ijd}^{l,m} - \hat{X}_{S, ijd}^{l,m})^2] },
\end{aligned}
\end{equation}
where $H, W, D$ denote the height,
width and the number of channels. The same equation can be applied on channel dimension with $\mathcal{M}=\{3\}$ to form $L_{ckd}$.
Overall,
the distillation objective can be denoted as:
\begin{equation}
    L = L_{skd} + L_{ckd}.
\end{equation}

\textbf{Average pooling\quad\vs\quad Strip pooling}. 
Benefiting from its strong ability to aggregate features and model long-range dependency,
strip pooling shines in many fully-supervised semantic segmentation models~\cite{huang2019ccnet,hou2020strip}.
The performance of continual segmentation is still much worse than that of fully-supervised segmentation.
In the scenario of continual segmentation,
there are often more noise or errors in the prediction results than fully-supervised segmentation.
Thus,
in the distillation process,
when using strip pooling to aggregate features,
this long-range dependency will introduce some uncorrelated noise to the cross point,
causing noise diffusion.
This will lead to further deterioration of the prediction results of the student model.
In our method,
we use average pooling in the local region to suppress the negative impact of noise.
Specifically,
because the semantics of local regions are often similar,
the current key point can find more neighbors to support its decision by aggregating features in the local region.
Thus, the current key point is less negatively affected by the noise in the local region.

\begin{figure}[t]
    \centering
	\begin{small}
		\begin{overpic}[width=0.47\textwidth,tics=8]{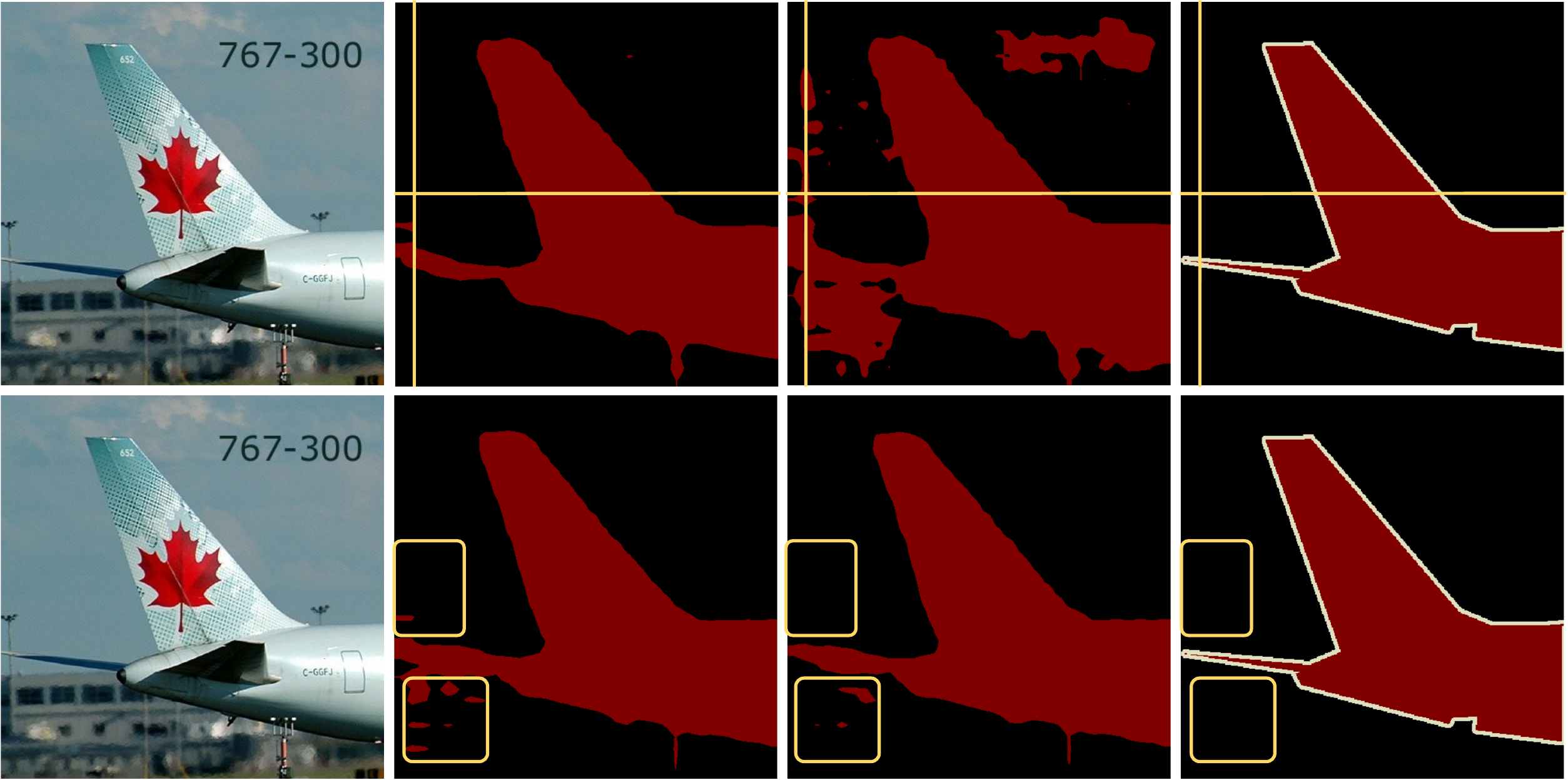}
	        \put(5, -3){(a) Image}
	        \put(30, -3){(b) step 2}
	        \put(55, -3) {(c) step 3}
	        \put(82, -3) {(d) GT}
		\end{overpic}
	\end{small}
	\vskip 0in
	\caption{
	The impact of the strip pooling (top row) used in PLOP~\cite{plop} and the average pooling (bottom row) in our method.
	}\label{fig:kd-motivation}
\end{figure}

As an example shown in~\figref{fig:kd-motivation}(b) top,
the strip pooling introduces noise or errors to the cross point for the teacher model.
During the distillation process,
the noise is further propagated to the student model,
making the noise diffusion.
For the average pooling in~\figref{fig:kd-motivation} bottom,
the key point will consider many nearby neighbors, resulting in an aggregated feature that is more robust to noise.

\section{Experiments}
In this section, we first demonstrate the details of our experimental setups, \eg{datasets, protocols and training details.}
Then we illustrate the effectiveness of our method from quantitative and qualitative experiments.

\subsection{Experimental setups}
\subsubsection{Datasets}

\textbf{PASCAL VOC 2012}~\cite{pascal-voc-2012}
is a commonly used dataset,
which contains 10,582 training images and 1449 validation images with 20 object classes and the background class.
\textbf{ADE20K}~\cite{zhou2017scene} is a dataset for semantic segmentation
covering daily life scenes.
It contains 20,210 training images and 2,000 validation images with 150 classes.
\textbf{Cityscapes}~\cite{cordts2016cityscapes}
contains 2,975 training images,
500 validation images and 1,525 test images.
There are 19 classes from 21 cities.

\subsubsection{Protocols}
\paragraph{Continual Class Segmentation.}

In continual class segmentation, the model is trained to recognize different classes sequentially in multiple steps. Each step the model learns one or several classes. Following~\cite{mib,plop,sdr}, we assume training data of previous steps are not available, \ie the model can only access data of the current step. Besides, only classes to be learned in the current step are labeled. All other classes are treated as \emph{background}.  
There are two commonly used settings proposed by~\cite{mib} for continual class segmentation,
\emph{disjoint} and \emph{overlapped}.
In the \emph{disjoint} setting,
assuming we know all classes in the future,
the images in the current training step do not contain any classes in the future.
The \emph{overlapped} setting is more realistic.
It allows potential classes in the future to appear in the current training images.

We conduct continual class segmentation experiments on the PASCAL VOC 2012~\cite{pascal-voc-2012}
and ADE20K~\cite{zhou2017scene}.
Following~\cite{mib,plop,sdr},
as defined in~\secref{sec:prelimi},
we call each training on the newly added dataset as a \emph{step}.
Formally,
$X\text{-}Y$ denotes the continual setting in our experiments,
where $X$ denotes the number of classes that we need to train in the first step.
In each subsequent learning step,
the newly added dataset contains $Y$ classes.
On PASCAL VOC 2012~\cite{pascal-voc-2012},
we conduct experiments on three settings,
15-5 (2 steps),
15-1 (6 steps) and 10-1 (11 steps).
For example,
15-1 denotes that we train the model on the initial 15 object classes in the first step.
In the subsequent five steps,
the model is expected to be trained on new datasets, where each dataset contains one new added class.
Thus, the model can discriminate 20 object classes in the last step.
On ADE20K~\cite{zhou2017scene},
we apply four settings,
\emph{100-50 (2 steps)},
\emph{50-50 (3 steps)},
\emph{100-10 (6 steps)},
and
\emph{100-5 (11 steps)}
.

\paragraph{Continual Domain Segmentation.} It is proposed by~\cite{plop}.
Different from continual class segmentation,
this setting is to deal with the domain shift phenomenon rather than integrating new classes.
In the real-world scene,
domain shift can also occur frequently.
We assume the classes in different domains are the same.
The training data of the old domain is not accessible when training on new domain data.
We conduct continual domain segmentation experiments on Cityscapes~\cite{cordts2016cityscapes}.
Following PLOP~\cite{plop},
we regard the training data in each city as a domain.
We also apply three settings, \emph{11-5 (3 steps)},
\emph{11-1 (11 steps)} and \emph{1-1 (21 steps)}.
In these experimental settings,
we use the same recording as the continual class segmentation,
but each step adds new domains (cities) instead of classes.

\input{tables/voc-v2}
\begin{figure*}[!tp]
	\centering
	\xdef\xfigwd{\textwidth}
	\begin{subfigure}{0.32\textwidth}
		\includegraphics[width=1\linewidth]{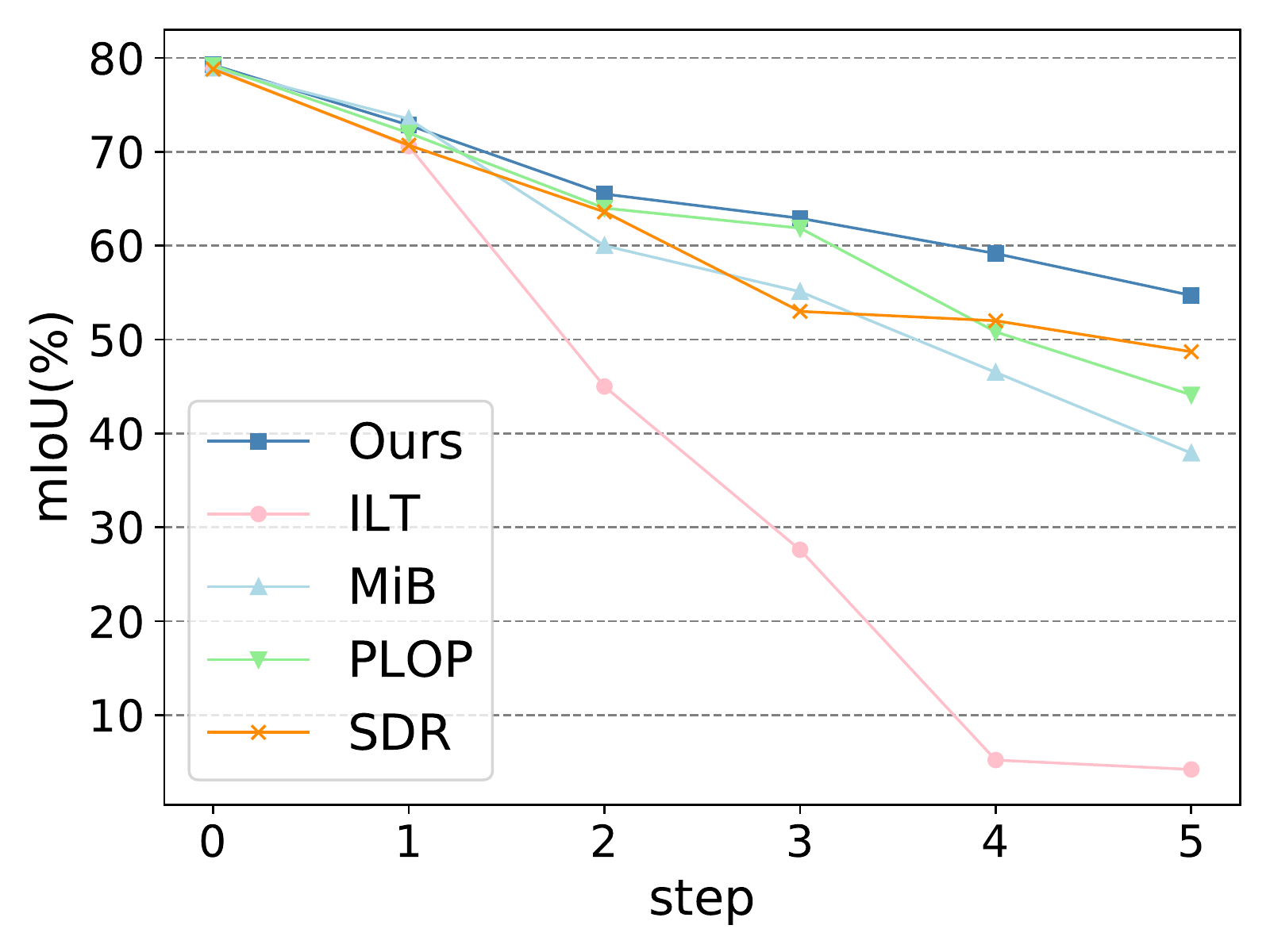}
		\caption{15-1 disjoint on PASCAL VOC 2012} \label{fig:all-disjoint}
	\end{subfigure}
	\begin{subfigure}{0.32\textwidth}
		\includegraphics[width=1\linewidth]{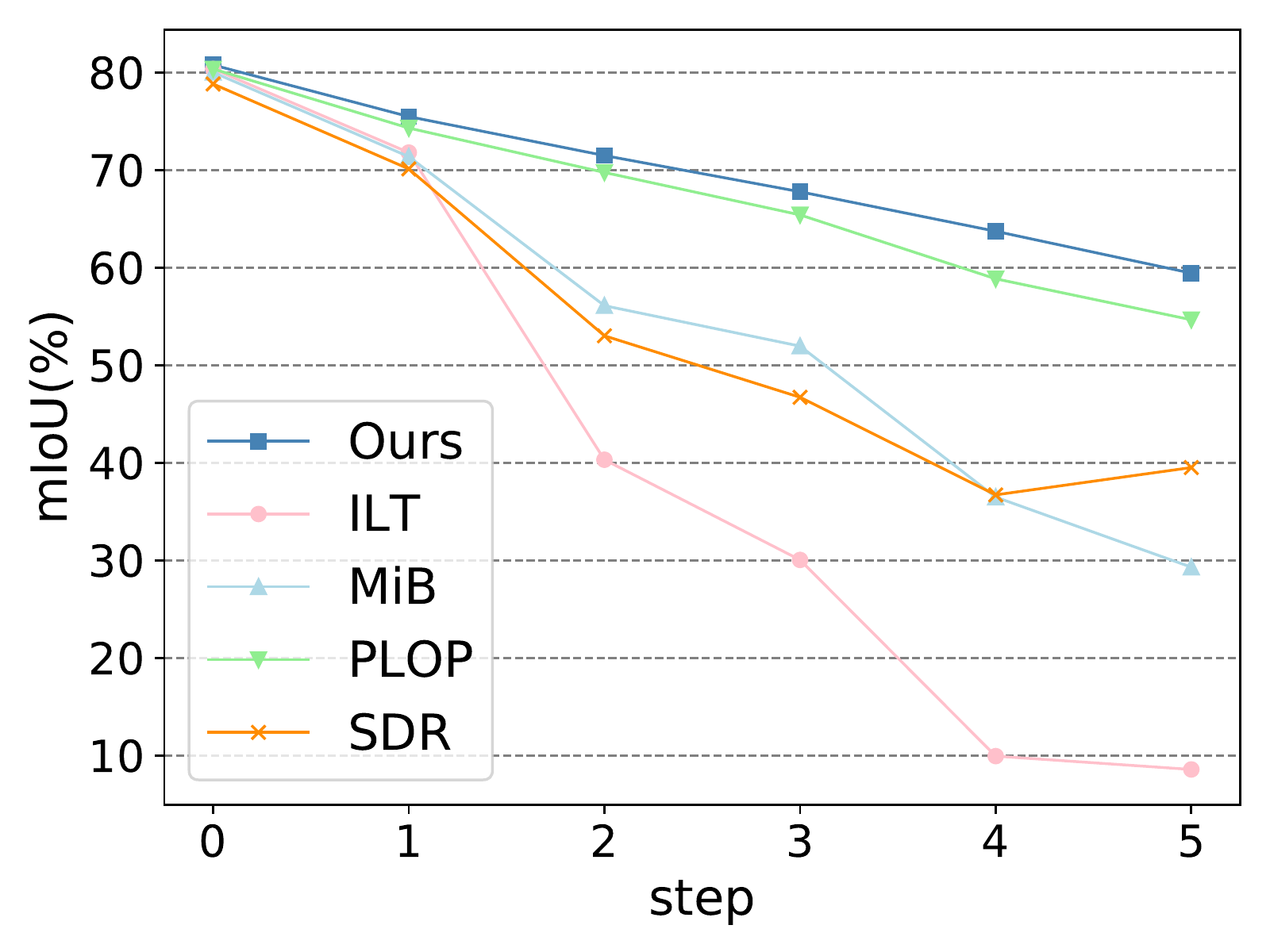}
		\caption{15-1 overlapped on PASCAL VOC 2012} \label{fig:all-overlapped}
	\end{subfigure}
	\begin{subfigure}{0.32\textwidth}
		\includegraphics[width=1\linewidth]{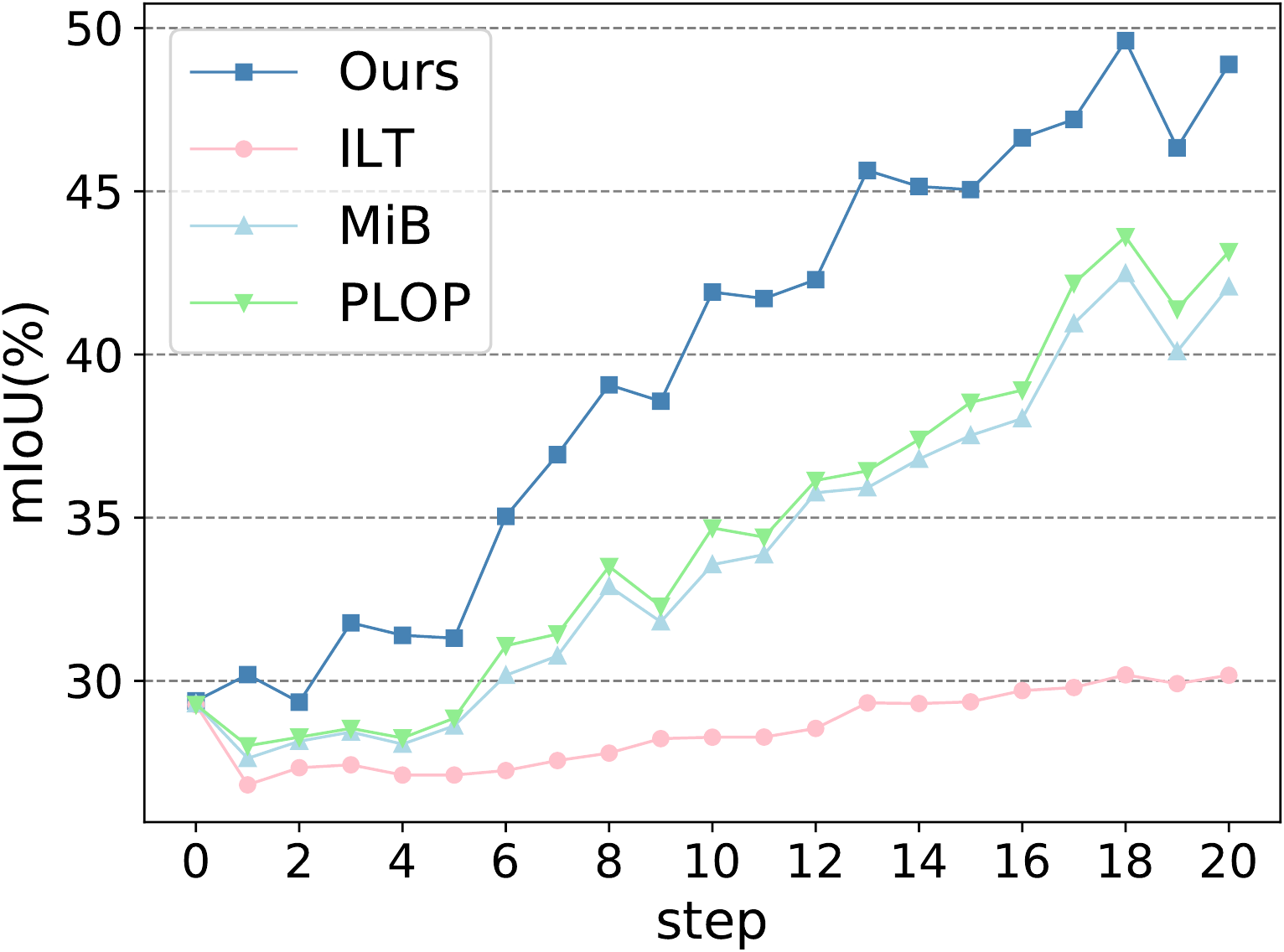}
		\caption{1-1 on Cityscapes} \label{fig:all-cityscapes}
	\end{subfigure}
	\caption{The mIoU (\%) at each step in three experimental settings. (a)(b) are settings of continual class segmentation. (c) is the setting of continual domain segmentation.}
\end{figure*}

\subsubsection{Implementation Details}
Following~\cite{mib,sdr,plop},
we use the Deeplab-v3~\cite{chen2017deeplab} architecture with ResNet-101\cite{he2016deep} as backbone.
The output stride of Deeplab-v3 is set to 16.
%
We also apply the in-place activated batch normalization~\cite{rota2018place} in the backbone pre-trained on the ImageNet~\cite{deng2009imagenet},
as the above methods.
We utilized the loss function proposed by MiB~\cite{mib} to assist our training process.
And we apply the same training strategy as~\cite{mib,plop,sdr}.
Specifically,
we apply the same data augmentation, \eg{horizontal flip and random crop}.
The batch size is set to 24 for all experiments.
We set the initial learning rate as 0.02 for the first training step and 0.001 for the next continual learning steps.
The learning rate is adjusted by the \emph{poly} schedule.
We train the model using SGD optimizer for each step with 30 (PASCAL VOC 2012~\cite{pascal-voc-2012}),
50 (Cityscapes~\cite{cordts2016cityscapes}),
and 60 epochs (ADE20K~\cite{zhou2017scene}), respectively.
We also use 20\% of the training set as validation following~\cite{mib,sdr,plop}.
We report the mean Intersect over Union (mIoU) on
the original validation set.
%

\subsection{Continual Class Segmentation}

\begin{figure*}[!htp] 
	\begin{small}
        \begin{overpic}[width=1\linewidth,tics=2]{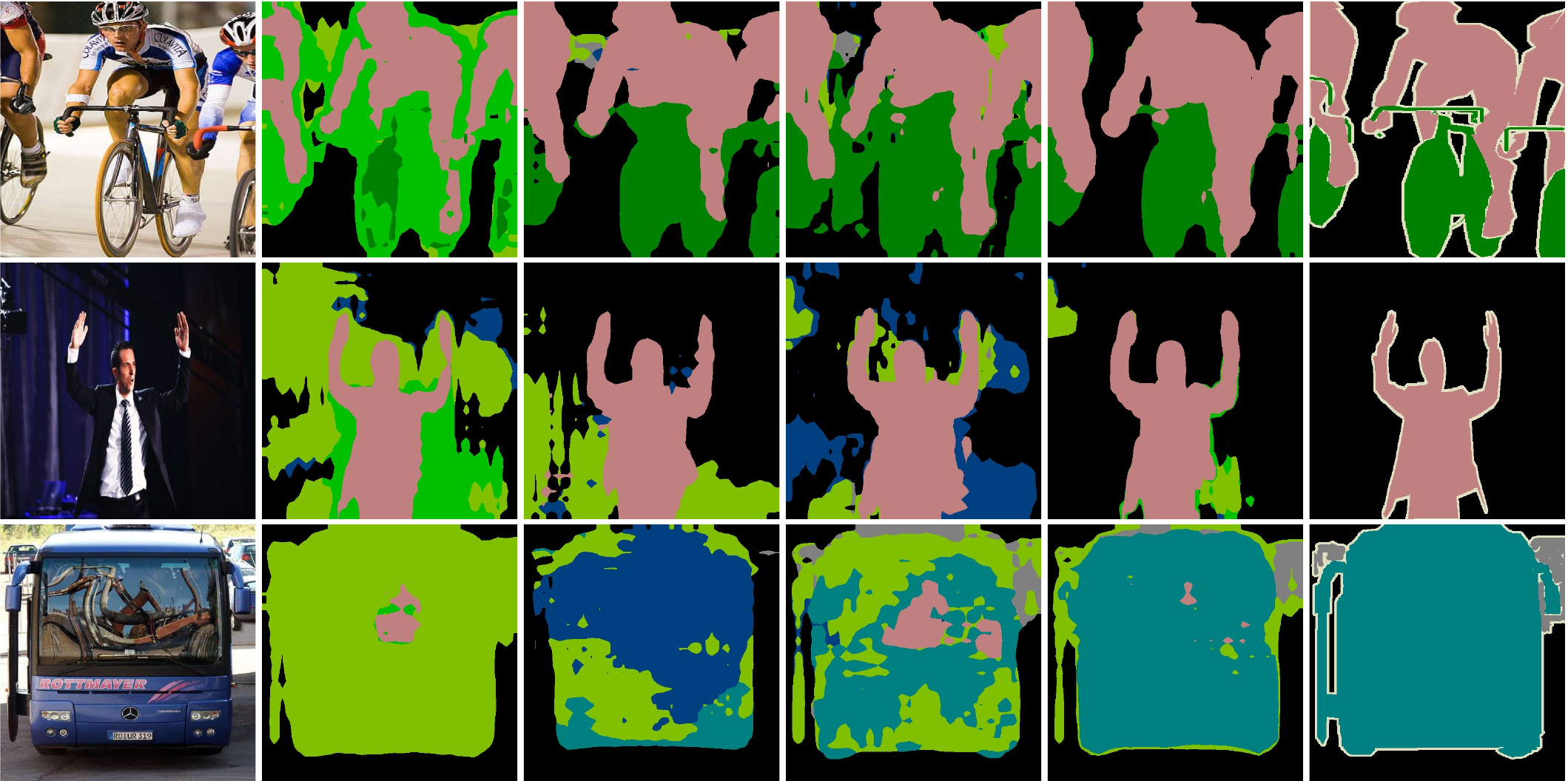}
            \put(6,-1.5) {Image}
            \put(22, -1.5){MiB~\cite{mib}}
            \put(38, -1.5){SDR~\cite{sdr}}
            \put(55, -1.5){PLOP~\cite{plop}}
            \put(73, -1.5){Ours}
            \put(90, -1.5){GT}
		\end{overpic}
	\end{small}
	\vskip -0.1in
	\caption{The qualitative comparison between different methods. All the prediction results are from the last step of 15-1 overlapped setting.
	}\label{fig:qualitative}
	\vskip -0.1in
\end{figure*}

\myPara{PASCAL VOC 2012.}
Applying the same experimental settings as~\cite{mib,plop,sdr},
we performed experiments on different continual learning settings, \emph{15-5},
\emph{15-1} and \emph{10-1}.
As shown in~\tabref{tab:pascal},
we report the experimental results of the last step.
The vanilla fine-tuning method suffers from the catastrophic forgetting phenomena.
The model quickly forgets the old knowledge and is unable to learn the new knowledge well.
Experimental results demonstrate that
our method significantly improves the segmentation performance
both on the overlapped and disjoint settings.
Especially in the challenging 15-1 settings,
our method outperforms the state-of-the-art by
6.0\% (\emph{disjoint}) and 4.8\% (\emph{overlapped}) in terms of mIoU, respectively.
We also display the performance of each step for different methods
as shown in~\figref{fig:all-disjoint} and \figref{fig:all-overlapped}.
This demonstrates that our method can reduce the forgetting of old knowledge in the continual learning process.
In~\tabref{tab:pascal},
we also report the performance over the old classes and new classes, respectively.
For all settings,
the performance of the old classes is greatly improved.
This is benefited from the representation compensation module and distillation mechanism,
which can effectively retain the old knowledge.
On the other hand,
our proposed representation module and distillation mechanism
allow room for learning new knowledge.
In~\secref{sec:ablation},
we will further analyze the effectiveness of these two mechanisms.
We further show the qualitative results of different methods
in the 15-1 \emph{overlapped} setting in~\figref{fig:qualitative}.

\input{tables/ade20k_overlap}
\input{tables/ade20k_100-5_overlap}

\paragraph{ADE20K.}
To verify the effectiveness of our method,
we conduct experiments on a challenging semantic segmentation dataset,
ADE20K~\cite{zhou2017scene}.
Experimental results are shown in~\tabref{tab:ade} and~\tabref{tab:ade20k-100-5}.
On different continual learning tasks, 100-50, 100-10 and 50-50,
our method achieves an average improvement of 1.4\% over the state-of-the-art.
To further verify our method,
we also perform experiments on a more challenging scenario, 100-5,
which contains 11 steps.
In this scenario,
our method also achieves the state-of-the-art, outperforming the previous method
by about 0.9\% in terms of mIoU,
as shown in~\tabref{tab:ade20k-100-5}.
The improvement is due to our proposed representation compensation module
and pooled cube distillation mechanism.

\subsection{Continual Domain Segmentation}
In the context of continual semantic segmentation,
in addition to the need to segment new classes,
it is also of great significance to increase the processing capabilities
of new domains.
Following~\cite{plop},
we conducted experiments of continual domain semantic segmentation
on Cityscapes~\cite{cordts2016cityscapes}.
Each city in Cityscapes~\cite{cordts2016cityscapes} can be regarded
as a domain, which is widely used by domain adaptive semantic segmentation
tasks~\cite{chen2017no}.
In this scenario,
we do not consider the difference in classes between domains.
As shown in~\tabref{tab:cityscapes},
experimental results demonstrate that
our method achieves higher mIoU than previous methods~\cite{ilt,mib,plop}
in all three settings.
Our method outperforms the state-of-the-art by 3.7\% on the challenging 1-1 setting
with 21 learning steps.
For this setting,
we display the performance of each step in~\figref{fig:all-cityscapes}.
Since MiB~\cite{mib} aims at solving the problem of semantic shift which is not existing in continual domain segmentation,
MiB~\cite{mib} performs slightly worse than Fine-tuning.
These experiments indicate that our method is also effective for continual domain semantic segmentation,
benefiting from the ability to retain old knowledge while allowing to learn new knowledge.

\input{tables/cityscapes}

\input{tables/ablation}
\subsection{Ablation Study} \label{sec:ablation}
In this section,
we firstly analyze the effectiveness of our proposed representation compensation
and pooled cube distillation mechanism.
Then we discuss the robustness to class orders in the continual learning scenario.

\textbf{Representation Compensation.}
We conduct ablation experiments on PASCAL VOC 2012~\cite{pascal-voc-2012}.
As shown in~\tabref{tab:ablation},
our proposed representation compensation module achieves about 7\% improvement than the MiB~\cite{mib} baseline.
With this module,
our method reaches state-of-the-art performance.
We argue this performance benefits from the scheme of remembering old knowledge in our method while allowing the learning for new knowledge.
In our method,
the operations of merging and freezing parameters aim at alleviating the forgetting of old knowledge.
Thus,
in~\tabref{tab:ablationRC},
we further study the effectiveness of these two operations.
Specifically, 
based on the plain parallel convolution branches (Parallel-Conv),
the operations of merging (Merge) and freezing (Frozen) can bring 2.7\% improvement. 
Experimental results demonstrate that
the model can benefit from the frozen knowledge in previous steps.

\input{tables/ablation-RC}
\input{tables/differentPooling}
\textbf{Distillation Mechanism.}
In~\tabref{tab:ablation},
we study the importance of knowledge distillation mechanism on spatial and channel dimensions, respectively.
The knowledge distillation on spatial and channel dimensions achieves similar performance,
outperforming baseline by about 15.3\% in terms of mIoU.
With the representation compensation module,
the combination of these two distillation schemes can
reach state-of-the-art performance.
We further compare the effectiveness of different pooling methods used in the knowledge distillation mechanism, as shown in~\tabref{tab:ablation-poolings}.
Experimental results demonstrate that average pooling outperforms strip pooling by 1.5\%.

\begin{figure}[!thp] 
	\centering
	\begin{small}
		\centering
		\begin{subfigure}{0.49\linewidth}
			\includegraphics[width=\linewidth]{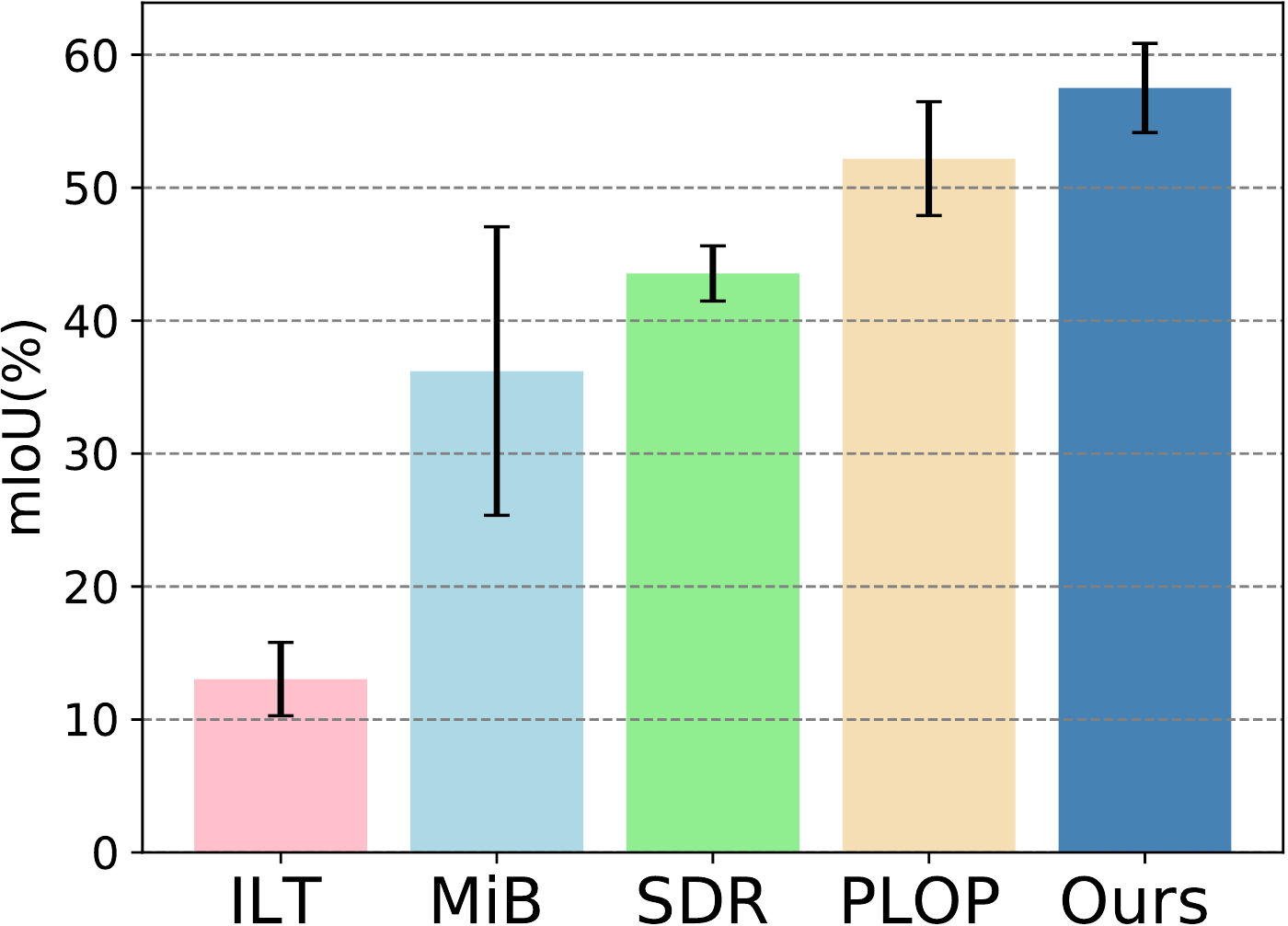}
			\caption{15-1 overlapped}
		\end{subfigure}
		\begin{subfigure}{0.49\linewidth}
			\includegraphics[width=\linewidth]{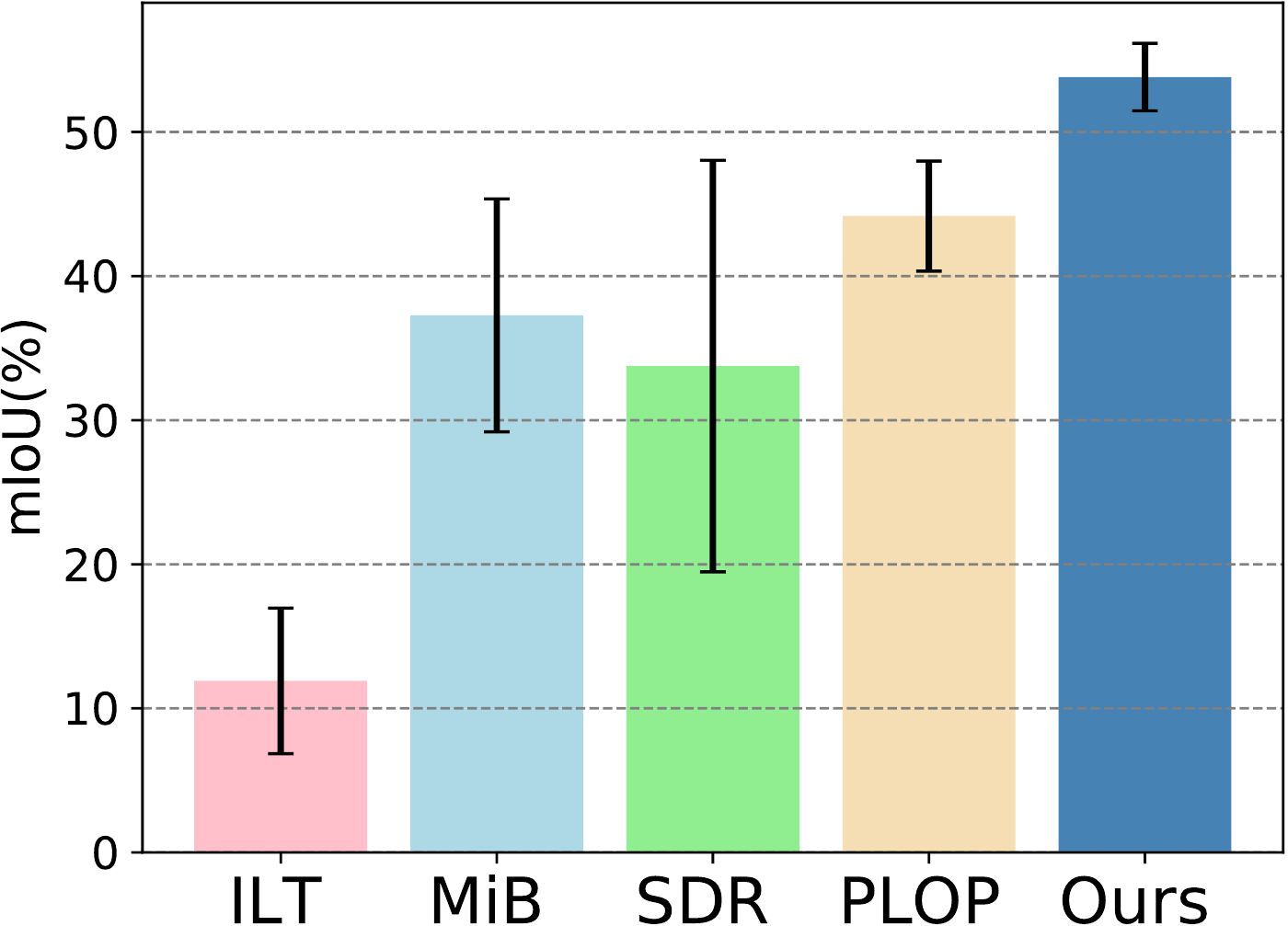}
			\caption{15-1 disjoint}
		\end{subfigure}
	\end{small}
	\vskip -0.1in
	\caption{The average performance and standard variance under different continual learning class orders.
	}\label{fig:classorder}
	\vskip -0.2in
\end{figure}

\textbf{Robustness to Class Orders.}
In the scenario of continual semantic segmentation, the class orders in the pipeline is particularly important.
To verify the robustness to class orders,
we perform experiments on five different class orders, including four random orders and the original ascending order.
In~\figref{fig:classorder},
we display the average performance and standard variance for different methods~\cite{ilt,mib,sdr,plop}.
Experimental results demonstrate that our method is more robust against different class orders than previous methods.

\section{Conclusion and Limitation}
In this work,
aiming at remembering the knowledge for old classes while allowing capacity for learning new classes,
we propose the representation compensation module,
which dynamically expands the network without any extra inference cost.
Besides,
to further alleviate the forgetting for old knowledge,
we propose Pooled Cube Distillation mechanism on spatial and channel dimensions.
We conduct experiments on two commonly used benchmarks,
continual class segmentation and continual domain segmentation.
Our method outperforms state-of-the-art performance.
%


Although we have proposed two components, which outperform the state-of-the-art performance,
we have a poor performance in the continual learning process with many steps, like 10-1 setting shown in~\tabref{tab:pascal}.
In these challenging scenarios,
how to improve the performance of the model still has a long way to go.
Besides,
our method requires more computation costs during training.

\textbf{Acknowledgment}
This work is funded by the National Key Research and Development Program of China (NO. 2018AAA0100400) ans NSFC (NO. 61922046), and S\&T innovation project from Chinese Ministry of Education.


{\small
\bibliographystyle{style/ieee_fullname}
\bibliography{ref}
}

\clearpage

\appendix
\section{Overview}
In~\secref{sec:rconcls},
We firstly conduct experiments on classification~\cite{wu2019large,douillard2020podnet} in continual learning.
We describe the details of our method in~\secref{sec:reproduce},
where we provide the pseudo-code for our proposed Pooled Cube Distillation.
Then we discuss some of the characters in our method in~\secref{sec:discussion},
including the robustness of our method against different class orders,
the impact of hyper-parameters  ,and
the ablation study about pooled knowledge distillation.
More importantly,
we explore our proposed representation compensation mechanism in~\secref{sec:explore-rc}.
Lastly,
we display more qualitative results in~\secref{sec:qualitive}.

\section{Continual Learning on Classification.}\label{sec:rconcls}
Our proposed representation compensation module can be easily integrated with many existing continual learning methods~\cite{wu2019large,douillard2020podnet,ahn2021ss}.
We conducted experiments integrating our representation compensation mechanism and two existing methods, EEIL~\cite{wu2019large} and PODNet~\cite{douillard2020podnet}.
We follow PODNet~\cite{douillard2020podnet} and report Top-1 accuracy on ImageNet-Subset (100 classes in total) with 50 classes as the first task and the rest are equally divided into five tasks (step 1 - 5).
As shown in Tab.~\ref{tab:classification}, For both EEIL and PODNet baselines, our method improves over them by about 2\% in average accuracy, respectively.

\begin{table}[!h]
    \centering
    \small
    \setlength{\tabcolsep}{2.5pt} 
    \begin{tabular}{c||c|c|c|c|c||c}
    \hline
        Method & step 1 & step 2 & step 3 & step 4 & step 5 & Avg Gain\\\hline 
        EEIL~\cite{wu2019large}  & 74.27 & 70.03 & 68.45 & 64.62 & 61.62 & \\
        + RC & \bf{77.23} & \bf{72.06} & \bf{70.10} & \bf{66.89} & \bf{63.82} & 2.22 \textcolor{red}{$\uparrow$} \\\hline 
        PODNet~\cite{douillard2020podnet} & 81.20 & 72.74 & 66.15 & 61.47 & 57.44 & \\
        + RC & \bf{81.90} & \bf{74.77} & \bf{70.05} & \bf{64.22} & \bf{60.04} & 2.40 \textcolor{red}{$\uparrow$}\\
    \hline
    \end{tabular}
    \caption{Experiments in Continual Classification. All experiments are conducted on the ImageNet-100.}
    \label{tab:classification}
\end{table}

\section{Reproducibility}\label{sec:reproducibility} \label{sec:reproduce}
In this section,
we describe more details about the loss functions.
Then we provide the pseudo-code for our proposed pooled cube distillation.

\begin{algorithm}[t]
\caption{Pseudo-code of Pooled Cube Distillation in a PyTorch-like style.}
\label{alg:code}
\algcomment{\fontsize{7.2pt}{0em}\selectfont \texttt{bmm}: batch matrix multiplication; \texttt{mm}: matrix multiplication; \texttt{cat}: concatenation.
}
\definecolor{codeblue}{rgb}{0.25,0.5,0.5}
\lstset{
  backgroundcolor=\color{white},
  basicstyle=\fontsize{7.2pt}{7.2pt}\ttfamily\selectfont,
  columns=fullflexible,
  breaklines=true,
  captionpos=b,
  commentstyle=\fontsize{7.2pt}{7.2pt}\color{codeblue},
  keywordstyle=\fontsize{7.2pt}{7.2pt},
}
\begin{lstlisting}[language=python]
# f_old: list of features from different stages of the old model
# f_new: list of features from different stages of the new model
# gamma: the hyper-parameters for the loss function
# loss_spatial: Pooled Cube distillation loss on spatial dimension
# loss_channel: Pooled Cube distillation loss on channel dimension

def pooledCube_KD(f_old, f_new):
    # define the average pooling size
    kernel_spatial = [4,8,12,16,20,24]
    kernel_channel = [3]
    
    # do PCD for different pairs of features
    for i, (x_old, x_new) in zip(f_old, f_new):
    
        #x_old: NxCxHxW
        PCD_old = hadamard_product(x_old, x_old) 
        #x_new: NxCxHxW
        PCD_new = hadamard_product(x_new, x_new) 
        
        loss_spatial = 0
        
        # multi-scale pooled cube distillation on spatial dimension
        for kernel in kernel_spatial:
            PCD_old = AvgPool2d(PCD_old, kernel)
            PCD_new = AvgPool2d(PCD_new, kernel)
            PCD_gap = (PCD_old - PCD_new).view(N,-1)
            PCD_gap = hadamard_product(PCD_gap, PCD_gap)
            
            loss_spatial += sqrt(PCD_gap.sum())
        
        loss_spatial /= len(kernel_spatial)
        
        # pooled cube distillation on channel dimension
        PCD_old = x_old.permute(0,2,1,3)
        PCD_new = x_new.permute(0,2,1,3)
        
        loss_channel = 0
        
        for kernel in kernel_channel:
            PCD_old = AvgPool2d(PCD_old, (kernel, 1))
            PCD_new = AvgPool2d(PCD_new, (kernel, 1))
            PCD_gap = (PCD_old - PCD_new).view(N,-1)
            PCD_gap = hadamard_product(PCD_gap, PCD_gap)
            
            loss_channel += sqrt(PCD_gap.sum())
        
        loss_channel /= len(kernel_channel)
        
    # compute the total loss
    loss = (loss_channel + loss_spatial) * gamma
    
    return loss

\end{lstlisting}
\end{algorithm}

\myPara{Objective.}

In the scenario of continual class semantic segmentation,
to save the labeling cost,
only the new classes are labeled in the newly added training data,
and the old classes are treated as the \emph{background} class.
Thus,
this brings a great challenge in continual class semantic segmentation, semantic shift~\cite{mib}.
To solve this issue,
we also apply the loss functions $L_{unce}$ and $L_{unkd}$ proposed by~\cite{mib} as~\cite{mib,sdr} in our pipeline as the baseline.
We refer to~\cite{mib} for more details.

Specifically,
let $C_t$ denotes the classes learned in step $t$.
Thus,
for the example $(x, y)$,
the objective for learning new classes can be written as
\begin{equation} \label{eq:celoss}
    L_{unce} = - \frac{1}{|\mathcal{I}|} \sum_{i\in \mathcal{I}} \log \hat{p_t}(i, y_i),
\end{equation}
where $y_i \in \{0, C_t\}$ denotes the ground-truth in the label for the $i$-th pixel.
And $\hat{p_t}(i)$ is modified from the predictions of current model $p_t(i)$, considering all old classes are \emph{background}.
The predicted scores for the old classes are summed to the \emph{backgroud} class.
The model is also supposed to maintain discrimination for old classes.
Thus,
the knowledge distillation objective can be denoted as
\begin{equation} \label{eq:unkdloss}
    L_{unkd} = - \frac{1}{|\mathcal{I}|} \sum_{k\in \mathcal{C}} \sum_{i\in \mathcal{I}} p_{t-1}(i, k) \log \hat{p_t}(i, k),
\end{equation}
where $p_{t-1}$ is the prediction of the old model, and $\mathcal{C}$ denotes all old classes and the \emph{background} class.
The $\hat{p_t}(i)$ is modified by predicted scores $p_t(i)$ of the current model.
In this objective,
all new classes are treated as the \emph{background} class,
and their predicted scores are summed to the \emph{background} class.

In this work,
the overall objective can be denoted as:
\begin{equation}
    L = L_{unce} + \lambda L_{unkd} \cdot \sqrt{\frac{||C||}{||C_t||}} + \gamma(L_{skd} + L_{ckd}),
\end{equation}
where $L_{skd}$ and $L_{ckd}$ denote the distillation loss function on spatial and channel dimensions, respectively.
The $\lambda, \gamma$ are hyper-parameters to balance the different objectives.
And the $||C||$ and $||C_t||$ denote the number of classes of all and current, respectively.
In our experiments,
we set the $\lambda$ as 100, and the $\gamma$ as 0.01.
We discuss the impact of hyper-parameters in~\secref{sec:discuss-hyper-param}.

\myPara{Pooled Cube Distillation.}
To further alleviate catastrophic forgetting,
we design pooled cube distillation strategy on both spatial and channel dimensions.
We display the pseudo-code in~\algref{alg:code}.

\section{Discussion}\label{sec:discussion}
\subsection{Robustness to Class Order}
To verify the impact of different class orders,
we run different methods on five different orders on the 15-1 overlapped setting,
which includes the ascending order and four random orders.
The four random orders are provided by the code of PLOP~\cite{plop}.
Experimental results are shown in~\tabref{tab:classorders-results}. We can observe that ILT~\cite{ilt}, MiB~\cite{mib}, and SDR~\cite{sdr} are less stable to different orders with large variance. PLOP~\cite{plop} improves over these methods by using multi-scale feature distillations. Thanks to the proposed mechanisms, Ours is much more robust to different orders and also obtains the best performance in terms of mIoU. 
The five orders are defined as:

    \begin{equation}
        \scriptsize
        \begin{aligned}
            &{A: \{ [0, 1, 2, 3, 4, 5, 6, 7, 8, 9, 10, 11, 12, 13, 14, 15], [16], [17],  [18], [19], [20] \},}\\
            &{B: \{ [0, 12, 9, 20, 7, 15, 8, 14, 16, 5, 19, 4, 1, 13, 2, 11], [17], [3], [6], [18], [10]\},}\\
            &{C: \{ [0, 13, 19, 15, 17, 9, 8, 5, 20, 4, 3, 10, 11, 18, 16, 7], [12], [14], [6], [1], [2] \}}\\
            &{D: \{ [0, 15, 3, 2, 12, 14, 18, 20, 16, 11, 1, 19, 8, 10, 7, 17], [6], [5], [13], [9], [4]\},}\\
            &{E: \{ [0, 7, 5, 3, 9, 13, 12, 14, 19, 10, 2, 1, 4, 16, 8, 17], [15], [18], [6], [11], [20]\}}.\\
        \end{aligned}
    \end{equation}

%
\input{tables/classorder}

\subsection{Impact of Hyper-parameters}\label{sec:discuss-hyper-param}
As described in~\secref{sec:reproducibility},
there are two hyper-parameters $\lambda$ and $\gamma$ in our objective.
We study the impact of these hyper-parameters in~\tabref{tab:supp-ablation-hyper}. Our method achieves the best performances when  $\gamma=0.005$ and $\gamma=0.01$, with the selected $\gamma$,  it can perform well within a relatively large range of $\lambda$ from 20 to 200.  
In our experiments,
considering $lambda$ is set as 100 in~\cite{mib},
thus we apply the same hyper-parameters as~\cite{mib}.
We set $\lambda$ as 100 and $\gamma$ as 0.01.

\begin{figure*}[!t] 
  \centering
  \small
  \begin{overpic}[width=0.95\linewidth]{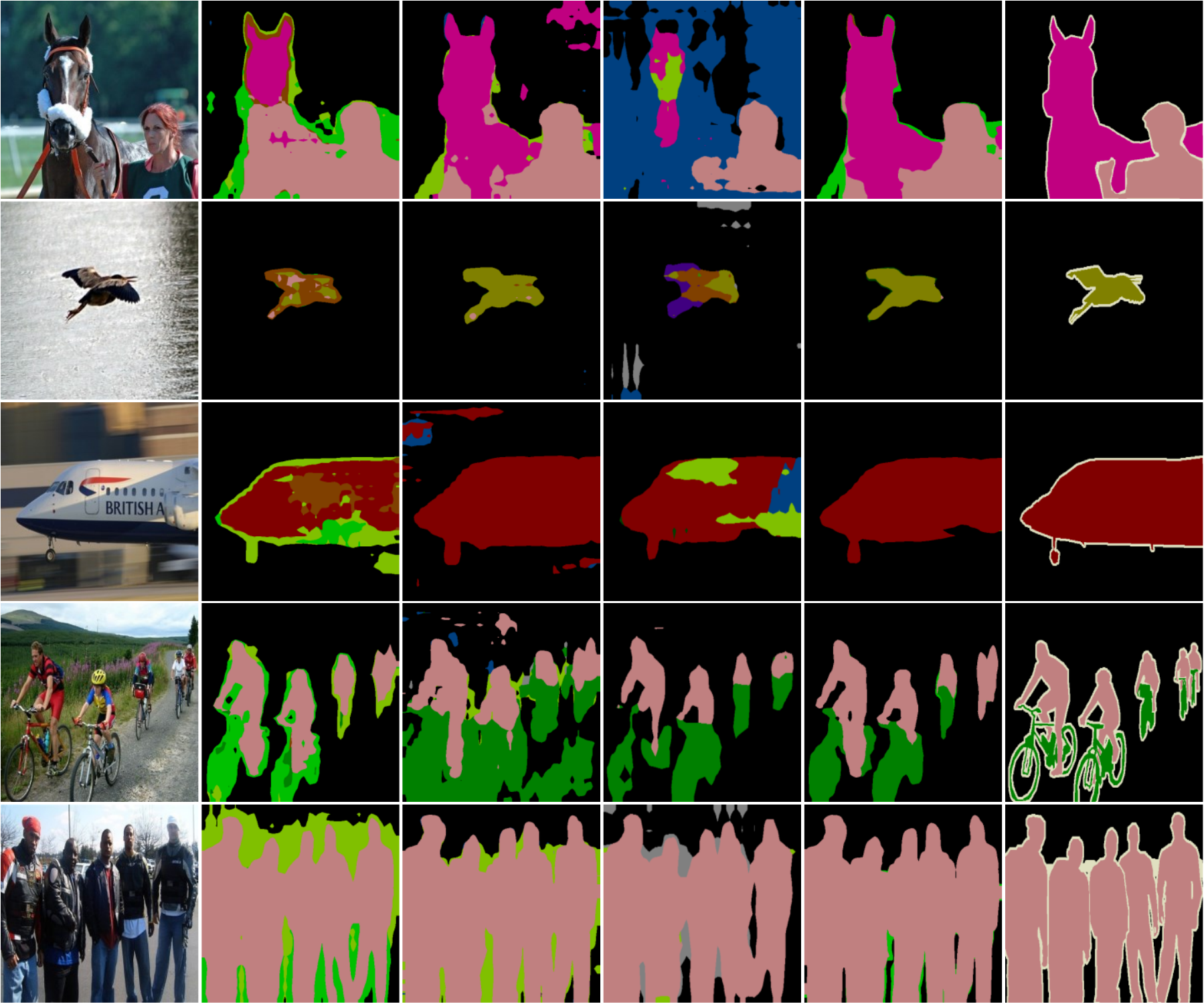}
    \put(6, -2) {Image}
    \put(23, -2){MiB~\cite{mib}}
    \put(39, -2){PLOP~\cite{plop}}
    \put(56.5, -2){SDR~\cite{sdr}}
    \put(73, -2){Ours}
    \put(90, -2){GT}
  \end{overpic}
  \vskip 0.1in
  \caption{Visualization results for different methods.
  }\label{fig:more-vis}
\end{figure*}

\begin{table}[!thp]
\begin{tabular}{c||c|c|c|c|c|c}
\toprule
\diagbox{$\lambda$}{$\gamma$}   & 0.0001 & 0.001 & 0.005 & 0.01 & 0.05 & 0.1 \\ \hline
1   &   35.4     &   39.8    &   46.3    &   49.3   &   46.5   &  42.8   \\
10  &   44.3     &   49.3    &   52.1    &   51.0   &   46.5   &  44.7   \\
20  &    49.0    &   56.9    &    57.6   &  56.1    &   50.0   &   47.8  \\
50  &    48.5    &   57.4    &   \textbf{59.7}    &  59.1    &   53.6   &   50.6  \\
100 &   42.9     &   55.0    &   \textbf{59.4}    & \underline{\textbf{59.4}}    &    55.5  &  50.8   \\
150 &    52.6    &   52.6    &   58.2    &  58.9    &   55.4   &  50.7   \\
200 &   50.0     &   50.0    &   57.8    &  58.3   &   55.1  &  51.0   \\ 
\bottomrule
\end{tabular}
\caption{Impact of different hyper-parameters. All experiments are conducted on the 15-1 overlapped setting on PASCAL VOC 2012 dataset. We select the $\lambda$ as 100 and $\gamma$ as 0.01 in our experiments.} \label{tab:supp-ablation-hyper}
\end{table}

\begin{table}[!tp]
\centering
\small
\setlength{\tabcolsep}{6pt} 
\begin{tabular}{c|c|c|c|c|c||c}
\toprule
4 & 8 & 12 & 16 & 20 & 24 & mIoU(\%) \\
\hline
\checkmark & & & & & &  55.1\\
 & \checkmark & & & & & \bf{56.2} \\
 &  & \checkmark & & & & \bf{56.2} \\
 &  & & \checkmark & & & 55.4 \\
 & & & & \checkmark & & 54.7 \\
 & & & & & \checkmark & 53.7 \\
\checkmark & \checkmark & & & & & 55.8 \\
\checkmark & \checkmark & \checkmark & & & & 56.1 \\
\checkmark & \checkmark & \checkmark & \checkmark & &  & \bf{56.2} \\
\checkmark & \checkmark & \checkmark & \checkmark & \checkmark &  & 56.1 \\
\checkmark & \checkmark & \checkmark & \checkmark & \checkmark & \checkmark & \underline{56.1}\\
\bottomrule
\end{tabular}
\vskip 0in
\caption{Impact of different average pooling kernel sizes in our proposed pooled cube distillation mechanism. All experiments are conducted on 15-1 overlapped on PASCAL VOC 2012 dataset using PLOP framework.} \label{tab:different-kernelsize}
\vskip -0.1in
\end{table}

\subsection{Ablation study about knowledge distillation}
\paragraph{Impact of different pooling kernel sizes.}
In the scenario of continual semantic segmentation,
the pooling operation plays a key role in the distillation mechanism.
In our distillation mechanism,
we use the multi-scale average pooling with kernel size in $\mathcal{M}=\{4,8,12,16,20,24\}$.
We study the impact of different pooling kernel sizes in~\tabref{tab:different-kernelsize}.
Experimental results demonstrate that if only one window size is used when the pooling window size is too small or too larger,
the performance will be worse.
We analyze that if the pooling window size is relatively small, when aggregating information for the current pixel, sufficient information of neighbors is not able to be  considered, so the negative impact of noise cannot be effectively suppressed.
When the pooling window size is relatively large, aggregating information for the current pixel can bring unrelated noise to the current pixel, therefore the performance is worse as well. When we combine multi-scale window sizes, the mIoU is stable at very high performance, therefore we use all scales as shown in ~\tabref{tab:different-kernelsize} in stead of choosing the optimal scales. 

\input{tables/kd-layers}
\paragraph{Distillation on different layers.}
We explore the impact of our proposed pooled cube distillation on different intermediate layers.
Experimental results are shown in~\tabref{tab:ablation-kdlayers},
which demonstrates that distillation on all layers outperforms the baseline without distillation by 21.7\% in terms of mIoU. It is interesting that distillation on the decoder gives the largest boost compared to other layers, which may be due to the high-level semantic information contained in the decoder. 
Thanks to deep supervision, the effect of gradient vanishing can be alleviated, and fusing distillation from all layers can further improve the performance. 
Therefore we use distillation on all layers in our work. 

\begin{figure}[!htp] 
  \centering
  \small
  \begin{overpic}[width=.95\linewidth]{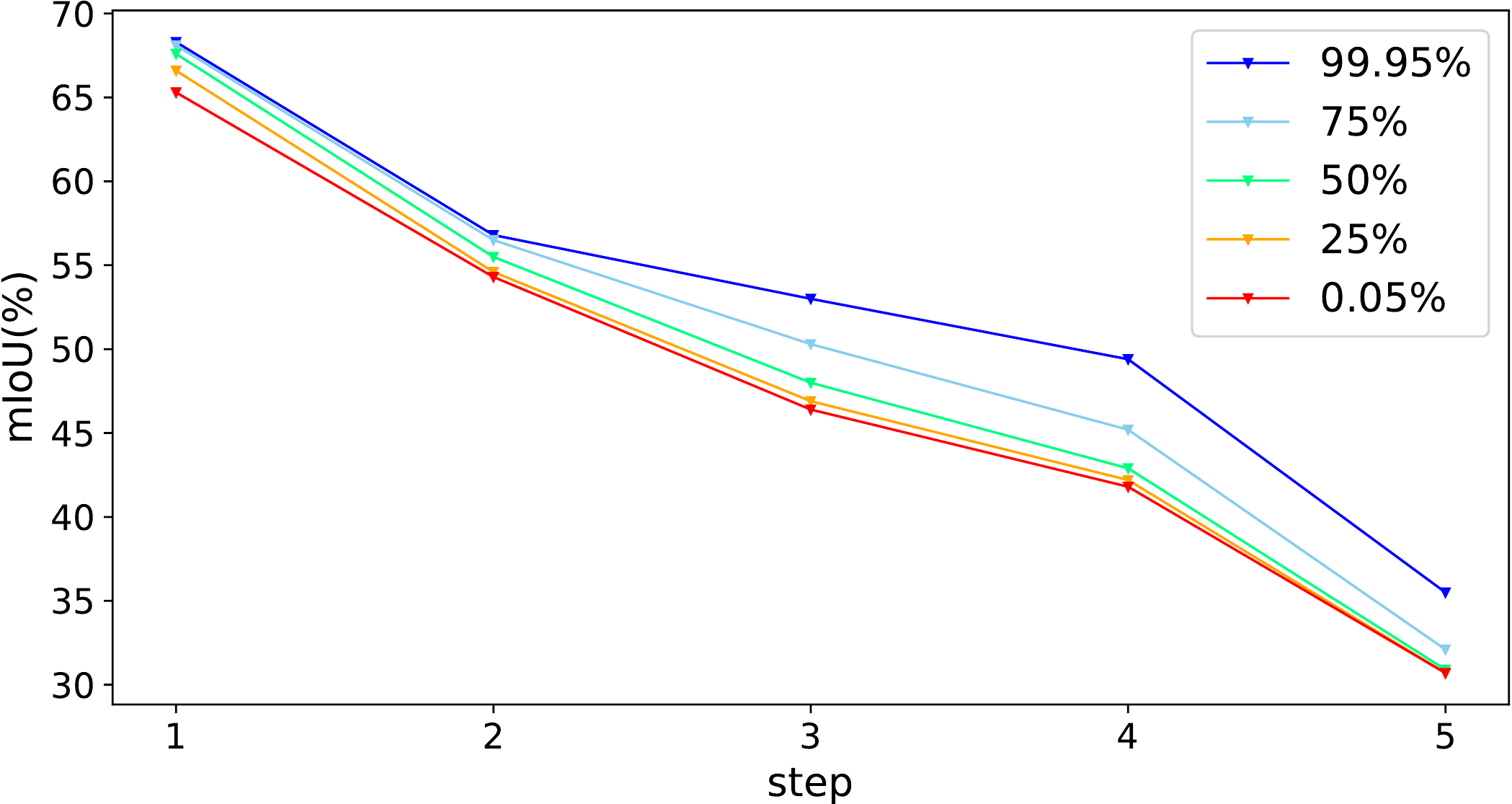}
  \end{overpic}
  \vskip -0.1in
  \caption{The mIoU(\%) for old classes.
  We set different weighting parameters (99.95\%, 75\%, 50\%, 25\%, 0.05\%) for the frozen branch during aggregating features from two branches. As the weight increases, the model presents a tendency to keep the memory of old knowledge. All experiments are conducted on 15-1 overlapped setting on PASCAL VOC 2012 dataset.
  }\label{fig:per-over-old}
  \vskip 0.2in
\end{figure}

\section{Exploring Representation Compensation} \label{sec:explore-rc}
Previously, we claimed that the left branch has the function of remembering the old knowledge, playing a role of great significance in preventing catastrophic forgetting. 
In~\figref{fig:per-over-old}, we let the left branch account for $99.95\%, 75\%, 50\%, 25\%, 0.05\%$ during fusion of training process to observe the model’s ability to remember old knowledge, \ie{the performance over old classes}. In order to ensure the fairness of the experiment and easy observation, we only added RC-module based on Fine-tuning. And we set the learning rate as 0.0001 for training from step 1 to step 5.
As shown in~\figref{fig:per-over-old}, with the weight increasing, the model gradually enhances the memory of old knowledge, indicating that the frozen branch can preserve the old knowledge.
Thus, our training process can benefit from this characteristic.

\section{More Qualitative Results} \label{sec:qualitive}
We display some visualization results in~\figref{fig:more-vis}.

\section{Future Work}
In our current RC-module,
the feature aggregation of two branches is obtained by linear weighting.
The weights indicate the importance of the two branches.
In our method,
we simply set the weights of two branches to 0.5.
We believe that it can achieve better performance by designing the feature aggregation method carefully.
For example,
in future work,
we could explore learnable weights for two branches.

\end{document}

%% file: tables/voc-v2.tex
\newcommand{\major}[1]{\textcolor{blue}{#1}}
\begin{table*}[!htp]
\centering
\setlength{\tabcolsep}{3pt} 
\small

\begin{tabular}{l||cc|c||cc|c||cc|c||cc|c||cc|c||cc|c}
\toprule
\multicolumn{1}{c}{}   & \multicolumn{6}{c}{{\textbf{15-5} (2 steps)}} & \multicolumn{6}{c}{{\textbf{15-1} (6 steps)}} & \multicolumn{6}{c}{{\textbf{10-1} (11 steps)}} \\
\multicolumn{1}{c||}{}    & \multicolumn{3}{c||}{\bf{Disjoint}}        & \multicolumn{3}{c||}{\bf{Overlapped}}  & \multicolumn{3}{c||}{\bf{Disjoint}}     & \multicolumn{3}{c||}{\bf{Overlapped}}  & \multicolumn{3}{c||}{\textbf{Disjoint}}      & \multicolumn{3}{c}{\textbf{Overlapped}} \\
\bf{Method} & \it{0-15}  & \it{16-20}   & \it{all}   & \it{0-15}  & \it{16-20}   & \it{all} & \it{0-15}  & \it{16-20}   & \it{all}  & \it{0-15}  & \it{16-20}   & \it{all} & \it{0-10} & \it{11-20} & \it{all} & \it{0-10} & \it{11-20} & \it{all}     \\ \hline
{Fine-tuning }     &  5.7    & 33.6    & 12.3      & 6.6  & 33.1  & 12.9    & 4.6  & 1.8  & 3.8  & 4.6        & 1.8   & 3.9 & 6.3 & 1.1 & 3.8 & 6.4 & 1.2 & 3.9      \\
{Joint} & 78.2&	78.0& 78.2	&	78.2&	78.0&	78.2&	79.8&	72.6&	78.2&	79.8&	72.6&	78.2 & 79.8       & 72.6       & 78.2       & 79.8       & 72.6       & 78.2 \\
\hline \hline
{LwF \cite{li2017learning} }    & 60.4  & 37.4 & 54.9 & 60.8  & 36.6  & 55.0  & 5.8 & 3.6 & 5.3 & 6.0 & 3.9 & 5.5 & 7.2 & 1.2 & 4.3 & 8.0 & 2.0 & 4.8 \\
{ILT \cite{ilt}}    & 64.9 & 39.5 & 58.9 & 67.8 & 40.6 & 61.3 & 8.6 & 5.7  & 7.9  & 9.6  & 7.8  & 9.2 & 7.3 & 3.2 & 5.4 & 7.2 & 3.7 & 5.5 \\
MiB~\cite{mib}   & 73.0  & 43.3    & 65.9  & 76.4   & 49.4  & 70.0  & 48.4   & 12.9   & 39.9    & 38.0   & 13.5       & 32.2 & 9.5 & 4.1 & 6.9 & 20.0 & 20.1 & 20.1    \\
SDR~\cite{sdr}& 74.6 & 44.1 & \textcolor{blue}{\bf{67.3}} & 76.3 & 50.2 & 70.1 & 59.4 & 14.3 & \textcolor{blue}{\bf{48.7}} & 47.3 & 14.7 & 39.5 & 17.3 & 11.0 & \textcolor{blue}{\bf{14.3}} & 32.4 & 17.1 & 25.1\\
PLOP~\cite{plop} & 71.0 & 42.8 & 64.3 & 75.7 & 51.7 & \textcolor{blue}{\bf{70.1}} &57.9 & 13.7 & 46.5 & 65.1 & 21.1 & \textcolor{blue}{\bf{54.6}} & 9.7 & 7.0 & 8.4 & 44.0 & 15.5 & \textcolor{blue}{\bf{30.5}} \\
\bf{Ours} & 75.0 & 42.8 & \textcolor{red}{\bf{67.3}} & 78.8 & 52.0 & \textcolor{red}{\bf{72.4}} & 66.1 & 18.2 & \textcolor{red}{\bf{54.7}} & 70.6 & 23.7 & \textcolor{red}{\bf{59.4}} & 30.6 & 4.7 & \textcolor{red}{\bf{18.2}} & 55.4 & 15.1 & \textcolor{red}{\bf{34.3}}\\

\bottomrule

\end{tabular}
\caption{The mIoU(\%) of the last step on the Pascal VOC 2012 dataset for different continual class segmentation scenarios. The \textcolor{red}{\bf{red}} denotes the highest results and the \textcolor{blue}{\bf{blue}} denotes the second highest results. }\label{tab:pascal}
\vskip 0.0in
\end{table*}

%% file: tables/ade20k_overlap.tex
\begin{table*}[!htp]
\centering
\small
\setlength{\tabcolsep}{3pt} 
\begin{tabular}{l||cc|c||cccccc|c||ccc|c}
\toprule
\multicolumn{1}{c}{} & \multicolumn{3}{c}{{\textbf{100-50} (2 steps)}} & \multicolumn{7}{c}{{\textbf{100-10} (6 steps)}} & \multicolumn{4}{c}{{\textbf{50-50} (3 steps)}} \\
\textbf{Method}       & \textit{1-100} & \textit{101-150} & \textit{all}  & \textit{1-100} & \textit{101-110} & \textit{111-120} & \textit{121-130} & \textit{131-140} & \textit{141-150} & \textit{all}  & \textit{1-50} & \textit{51-100} & \textit{101-150} & \textit{all}  \\ \hline
ILT~\cite{ilt} & 18.3 & 14.8 & 17.0 & 0.1 & 0.0 & 0.1 & 0.9 & 4.1 & 9.3 & 1.1 & 13.6 & 12.3 & 0.0 & 9.7\\
MiB~\cite{mib} & 40.7 & 17.7 & 32.8 &38.3 &12.6 &10.6 &8.7 &9.5 & 15.1 &29.2 & 45.3 & 26.1 & 17.1 & 29.3\\
PLOP~\cite{plop}   & 41.9 & 14.9 & \textcolor{blue}{\bf{32.9}} & 40.6 & 15.2 & 16.9 & 18.7 & 11.9 & 7.9 & \textcolor{blue}{\bf{31.6}} & 48.6 & 30.0 & 13.1 & \textcolor{blue}{\bf{30.4}}\\
\textbf{Ours}   & 42.3  & 18.8    & \textcolor{red}{\textbf{34.5}} & 39.3 & 14.6 & 26.3   & 23.2   & 12.1   & 11.8   & \textcolor{red}{\textbf{32.1}}   & 48.3 & 31.3   & 18.7    & \textcolor{red}{\textbf{32.5}} \\ \hline
Joint & 44.3  & 28.2    & 38.9 & 44.3  & 26.1   & 42.8   & 26.7   & 28.1   & 17.3   & 38.9 & 51.1 & 38.3   & 28.2    & 38.9\\
\bottomrule
\end{tabular}
\vskip 0in
\caption{The mIoU(\%) of the last step on the ADE20K dataset for different overlapped continual learning scenarios.
The \textcolor{red}{\bf{red}} denotes the highest results and the \textcolor{blue}{\bf{blue}} denotes the second highest results.}\label{tab:ade}
\vskip -0.1in
\end{table*}

%% file: tables/ade20k_100-5_overlap.tex
\begin{table}[!htp]
\centering
\small
\setlength{\tabcolsep}{6pt} 
\begin{tabular}{l|c||c||c}
\toprule

\textbf{Method}   & \textit{1-100} & \textit{101-150} & \textit{all} \\ \hline
ILT~\cite{ilt}   & 0.1  & 1.3 & 0.5 \\
MiB~\cite{mib}   & 36.0 & 5.6 & 25.9 \\
PLOP~\cite{plop} & 39.1 & 7.8 & \textcolor{blue}{\bf{28.7}} \\
Ours & 38.5 & 11.5 & \textcolor{red}{\bf{29.6}} \\
\bottomrule
\end{tabular}
\caption{The final mIoU(\%) of 100-5 overlapped on ADE20K.}\label{tab:ade20k-100-5}
\end{table}

%% file: tables/cityscapes.tex
\begin{table}[t]
\centering
\small
\setlength{\tabcolsep}{1pt} 
\begin{tabular}{l|c||c||c}
\toprule
\textbf{Method}       & \textbf{11-5} (3 steps) & \textbf{11-1} (11 steps) & \textbf{1-1} (21 steps)  \\ \hline
Fine-tuning    & 61.7  & 60.4  & 42.9 \\
LwF~\cite{li2017learning} & 59.7 & 57.3 & 33.0 \\
LwF-MC~\cite{rebuffi2017icarl} & 58.7 & 57.0 & 31.4 \\
ILT~\cite{ilt}  & 59.1 & 57.8 & 30.1 \\
MiB~\cite{mib}   & 61.5 & 60.0 & 42.2 \\
PLOP~\cite{plop}  & \textcolor{blue}{\bf{63.5}} & \textcolor{blue}{\bf{62.1}} & \textcolor{blue}{\bf{45.2}} \\
Ours & \textcolor{red}{\bf{64.3}} & \textcolor{red}{\bf{63.0}} & \textcolor{red}{\bf{48.9}} \\
\bottomrule
\end{tabular}
\caption{The final mIoU(\%) for continual domain semantic segmentation on Cityscapes~\cite{cordts2016cityscapes}.}\label{tab:cityscapes}
\end{table}

%% file: tables/ablation.tex
\begin{table}[!htp]
\centering
\small
\setlength{\tabcolsep}{4pt} 
\begin{tabular}{c|c|c|c|c||c}
\toprule
MiB$^\ddag$\cite{mib} & RC & Strip~\cite{hou2020strip} & S-KD & C-KD & 15-1 \\
\hline
 \checkmark & & & & & 36.1\\
 \checkmark & \checkmark& & & & 43.0 \\
 \checkmark & \checkmark & & \checkmark & & 58.3 \\
 \checkmark & \checkmark & & & \checkmark & 58.4 \\
\checkmark &  & & \checkmark & \checkmark & 57.8 \\
\checkmark & \checkmark & \checkmark & & & 57.9\\
 \checkmark & \checkmark & & \checkmark & \checkmark & \textbf{59.4} \\
\bottomrule
\end{tabular}
\vskip 0in
\caption{The final mIoU(\%) of ablation study about representation compensation module (RC) and pooled cube distillation mechanism on spatial (S-KD) and channel dimension (C-KD).
Experiments are conducted on 15-1 overlapped setting on PASCAL VOC 2012. $\dag$ denotes that the baseline is improved by an adaptive factor~\cite{plop}.}\label{tab:ablation}
\end{table}

%% file: tables/ablation-RC.tex
\begin{table}[!tp]
\centering
\small
\setlength{\tabcolsep}{6pt} 

\begin{tabular}{c|c|c|c||c}
\toprule
Parallel-Conv & Merge & Frozen & Drop-path & 15-1 \\
\hline
\checkmark &  & & & 40.1 \\
\checkmark & \checkmark & & & 42.0 \\
\checkmark & \checkmark & \checkmark & & 42.8 \\
\checkmark & \checkmark & \checkmark & \checkmark & \textbf{43.0} \\

\bottomrule
\end{tabular}
\vskip 0in
\caption{Ablation study of representation compensation module.
All experiments are conducted on PASCAL VOC 2012 without pooled cube distillation.}\label{tab:ablationRC}
\end{table}

%% file: tables/differentPooling.tex
\begin{table}[!tp]
\centering
\small
\setlength{\tabcolsep}{2pt} 
\begin{tabular}{c|c|c|c|c}
\toprule
W/o Pooling & GAP & Max Pooling & Strip Pooling & Avg. Pooling \\
\hline
52.0 & 36.1 & 48.0 & 54.6 & \bf{56.1} \\

\bottomrule
\end{tabular}
\vskip 0.in
\caption{Comparison between different pooling methods in distillation mechanism. All experiments are conducted on 15-1 overlapped on PASCAL VOC 2012 using PLOP framework. GAP denotes the global average pooling.}\label{tab:ablation-poolings}
\vskip -0.2in
\end{table}

%% file: tables/classorder.tex
\begin{table}[!htp]
\begin{tabular}{c|c|c|c}
\hline
Method                & Task & overlapped & disjoint \\ \toprule
\multirow{5}{*}{ILT~\cite{ilt}}  & A    & 9.20       & 7.90     \\
                      & B    & 16.74      & 20.65    \\
                      & C    & 12.16      & 6.37     \\
                      & D    & 11.49      & 10.85    \\
                      & E    & 15.60      & 13.77    \\
                      &   & \cellcolor{blue!80!gray!20!} 13.04 $\pm$ 2.76  & \cellcolor{blue!80!gray!20!} 11.91$\pm$5.05 \\ \hline \hline
\multirow{5}{*}{MiB~\cite{mib}}  & A    & 32.20      & 39.9     \\
                      & B    & 20.15      & 23.68    \\
                      & C    & 36.05      & 34.25    \\
                      & D    & 38.91      & 40.55    \\
                      & E    & 53.73      & 48.01    \\ 
                      &   & \cellcolor{blue!80!gray!20!} 36.21$\pm$10.8  & \cellcolor{blue!80!gray!20!}  37.28$\pm$8.08       \\\hline \hline
\multirow{5}{*}{SDR~\cite{sdr}}  & A    & 44.39      & 45.68    \\
                      & B    & 40.65      & 6.60     \\
                      & C    & 46.36      & 34.31    \\
                      & D    & 44.61      & 37.04    \\
                      & E    & 41.72      & 45.15    \\ 
                      &   & \cellcolor{blue!80!gray!20!} 43.55 $\pm$ 2.07  & \cellcolor{blue!80!gray!20!} 33.76 $\pm$ 14.29 \\    \hline \hline
\multirow{5}{*}{PLOP~\cite{plop}} & A    & 54.60      & 46.50    \\
                      & B    & 47.43      & 41.67    \\
                      & C    & 53.43      & 48.00    \\
                      & D    & 58.25      & 46.81    \\
                      & E    & 47.20      & 37.86    \\ 
                      &  & \cellcolor{blue!80!gray!20!} 52.18 $\pm$ 4.28  & \cellcolor{blue!80!gray!20!} 44.17 $\pm$ 3.82 \\     \hline \hline
\multirow{5}{*}{Ours} & A    & 59.40      & 54.70    \\
                      & B    & 54.05      & 53.26    \\
                      & C    & 55.63      & 49.53    \\
                      & D    & 55.29      & 55.53    \\
                      & E    & 63.19      & 56.07    \\ 
                      &   & \cellcolor{blue!80!gray!20!} 57.51 $\pm$ 3.35  & \cellcolor{blue!80!gray!20!} 53.82 $\pm$ 2.34 \\    \bottomrule
\end{tabular}
\caption{The mIoU(\%) of the final step. We conduct experiments on different class orders on 15-1 overlapped. The \colorbox{blue!80!gray!20!}{purple} denotes the mean mIoU(\%) and standard variance over five different class orders. } \label{tab:classorders-results}

\end{table}

%% file: tables/kd-layers.tex
\begin{table}[!tp]
\centering
\small
\setlength{\tabcolsep}{2pt} 

\begin{tabular}{c|c|c|c|c||c}
\toprule
\emph{layer 1} & \emph{layer 2} & \emph{layer 3} & \emph{layer 4} & \emph{decoder} & 15-1 \\
\hline
& & & & & 36.1 \\
\checkmark & & & & & 33.6 \\
 & \checkmark & & & & 34.0 \\
 & & \checkmark & & & 39.7 \\
 & & & \checkmark & & 47.2 \\
 & & & & \checkmark & 54.1 \\
\checkmark & \checkmark & & & & 32.8 \\
\checkmark & \checkmark & \checkmark & & & 34.0 \\
\checkmark & \checkmark & \checkmark & \checkmark & & 46.6\\
 &  &  & \checkmark & \checkmark & 55.3\\
 &  & \checkmark & \checkmark & \checkmark & 56.6\\
 & \checkmark & \checkmark & \checkmark & \checkmark & 57.4\\
\checkmark & \checkmark & \checkmark & \checkmark & \checkmark & \textbf{57.8} \\

\bottomrule
\end{tabular}
\vskip 0in
\caption{Ablation study about distillation mechanism at different stages.
	All experiments are conducted on the 15-1 overlapped setting on PASCAL VOC 2012 without  RC module.}\label{tab:ablation-kdlayers}
\vskip 0.1in
\end{table}